\documentclass[twocolumn,aps,longbibliography,superscriptaddress]{revtex4-1}
\usepackage{graphicx}  
\usepackage{dcolumn}   
\usepackage{bm}        
\usepackage{amssymb}   
\usepackage{textcomp}
\usepackage{mathtools}
\usepackage{color}
\usepackage{subfigure}
\usepackage{url}
\usepackage{cprotect}   
\usepackage[english]{babel}

\newcommand{\units}[1]{\; \text{#1}}
\newcommand{\e}[1]{\times 10^{#1}}

\definecolor{rev1}{rgb}{0,0,0}
\definecolor{rev2}{rgb}{0,0,0}
\definecolor{revs}{rgb}{0,0,0}
\newcommand{\revs}[1]{\textcolor{revs}{#1}}
\newcommand{\revv}[1]{\textcolor{rev2}{#1}}

\begin{document}

\title{Brain-inspired photonic signal processor for periodic \\ pattern generation and chaotic system emulation}

\author{Piotr Antonik}\thanks{piotr.antonik@ulb.ac.be}
\affiliation{Laboratoire d'Information Quantique, Universit\'e libre de Bruxelles, 50 Avenue F. D. Roosevelt, CP 224, B-1050 Brussels, Belgium}
\author{Marc Haelterman}\affiliation{Service OPERA-Photonique, Universit\'e libre de Bruxelles, 50 Avenue F. D. Roosevelt, CP 194/5, B-1050 Brussels, Belgium}
\author{Serge Massar$^1$}

\begin{abstract}
  Reservoir computing is a bio-inspired computing paradigm for processing time-dependent signals. Its hardware implementations have received much attention because of their simplicity and remarkable performance on a series of benchmark tasks. In previous experiments the output was uncoupled from the system and in most cases simply computed offline on a post-processing computer. However, numerical investigations have shown that feeding the output back into the reservoir would open the possibility of long-horizon time series forecasting. Here we present a photonic reservoir computer with output feedback, and demonstrate its capacity to generate periodic time series and to emulate chaotic systems. We study in detail the effect of experimental noise on system performance. In the case of chaotic systems, this leads us to introduce several metrics, \textcolor{rev1}{based on standard signal processing techniques,} to evaluate the quality of the emulation. Our work significantly enlarges the range of tasks that can be solved by hardware reservoir computers, and therefore the range of applications they could potentially tackle. It also raises novel questions in nonlinear dynamics and chaos theory.
\end{abstract}

\maketitle

\section{Introduction}

Reservoir Computing (RC) is a set of machine learning methods for designing and training artificial neural networks, introduced independently in \cite{jaeger2004harnessing} and in \cite{maass2002real}. The idea behind these techniques is that one can exploit the dynamics of a recurrent nonlinear network to process time series without training the network itself, but simply adding a general linear readout layer and only training the latter. This results in a system that is significantly easier to train (the learning is reduced to solving a system of linear equations \cite{lukovsevivcius2009survey}), yet powerful enough to match other algorithms on a series of benchmark tasks. RC has been successfully applied to, for instance, channel equalisation \cite{jaeger2004harnessing}, phoneme recognition \cite{triefenbach2010phoneme} and won an international competition on prediction of future evolution of financial time series \cite{NFC}.

Reservoir Computing allows to efficiently implement simplified recurrent neural networks in hardware, such as e.g. optical components. Optical computing has been investigated for decades as photons propagate faster than electrons, without generating heat or magnetic interference, and thus promise higher bandwidth than conventional computers \cite{arsenault2012optical}. RC would thus allow to build high-speed and energy efficient photonic \revs{computing} devices. Several important steps have been taken towards this goal with electronic \cite{appeltant2011information}, opto-electronic \cite{paquot2012optoelectronic,larger2012photonic,martinenghi2012photonic,larger2017high}, all-optical \cite{duport2012all,brunner2012parallel,vinckier2015high} and integrated \cite{vandoorne2014experimental} experimental RC implementation reported since 2012.

Forecasting is one of the central problems in science: how can we predict the future from the past? Over the past few decades, artificial neural networks have gained a significant popularity in the time series forecasting community. Compared to previously employed statistics-based techniques, they are both data driven and non-linear, more flexible and do not require an explicit model of the underlying process. A review of artificial neural networks models for time series forecasting can be found in \cite{zhang2012neural}.
Reservoir computing can be directly applied to short-term prediction tasks, that focus on generating a few future timesteps. 
As for long-horizon forecasting, that involves predicting the time series for as long as possible, 
it is possible with a small modification of the architecture, namely by feeding the RC output signal back into the reservoir.
This additional feedback significantly enriches the internal dynamics of the system, enabling it to generate time series autonomously, that is without receiving any input signal. With this modification, reservoir computing can be used for long-term prediction of chaotic series 
\cite{wyffels2010comparative,antonik2016towards,xu2016lnorm,NFC,jaeger2004harnessing}.
In fact, this approach holds, to the best of our knowledge, the record for such chaotic time series prediction \cite{jaeger2004harnessing, NFC}.
A reservoir computer with output feedback can also achieve the easier task of generating periodic signals \cite{wyffels2008stable,caluwaerts2013locomotion,reinhart2012regularization}, and of producing a tunable frequency \cite{wyffels2014frequency,antonik2016towards2,jaeger2007echo}.

The aim of the present work is to explore these novel applications experimentally. Indeed, they have been widely studied numerically, but no experimental implementation has been reported so far. There are multiple motivations for this investigation. First of all, reservoir computing is a biologically inspired algorithm. Indeed one of the main motivations of the seminal paper \cite{maass2002real} was to propose how microcircuits in the neocortex could process information. More recently it has been realised that the cerebellum has a structure very similar to that of a reservoir computer \cite{yamazaki2007cerebellum,rossert2015edge}. Generating time series with specific attributes is an important property of biological neural and chemical circuits (for e.g. movement control, biological rhythms, etc). Are biological circuits similar to reservoir computers used to generate trainable time series? Investigating this process experimentally can shed light on this tantalising question, for instance by clarifying which kinds of time series, and what training processes are robust to experimental imperfections.

Second, generation of time series with specific properties is an important task in signal generation and processing. Given the possibility that photonic reservoir computing could carry out ultra-fast and low energy optical signal processing, this is an important area to explore, again with the aim of understanding which tasks are robust to experimental imperfections.

Finally, this investigation raises a new fundamental question in nonlinear dynamics: given a system that emulates a known chaotic time series, how does one quantify the quality of the emulation. Answering this question becomes vital in case of experimental implementations, as physical systems are affected by noise, and thus can not output anything better than an approximate, noisy emulation of the target chaotic time series. \revs{The comparison techniques previously used in numerical investigations} fail in such situations, and one needs to develop new evaluation metrics.

Experimental implementation of these ideas requires, in principle, a fast readout layer capable of generating and feeding back the output signal in real-time. Several analogue solutions have been under investigation recently \cite{smerieri2012analog,duport2016fully,vinckier2016autonomous}, but none are as yet capable of realising this application. In fact, 
to successfully train an analogue readout layer with offline learning methods, used in most experimental RC setups up to now, a very precise model of the readout setup is required, which is hardly achievable experimentally, as shown in \cite{duport2016fully}: it is virtually impossible to characterise each hardware component of the setup with sufficient level of accuracy.
The reason for this sensitivity is that the output is a weighted sum with positive and negative coefficients of the internal states of the reservoir. Therefore errors in the coefficients quickly build up and become comparable to the value of the desired output.
For this reason, we chose the approach of a fast real-time digital readout layer implemented on a Field-Programmable Gate Array (FPGA) chip. The use of high-speed dedicated electronics makes it possible to compute the output signal in real time and feed it back into the reservoir. In order to keep the experiment simple, we used as reservoir the opto-electronic delay system introduced in \cite{appeltant2011information,paquot2012optoelectronic,larger2012photonic}, that has shown state-of-the-art results on several benchmark tasks and is fairly easy to operate. \revs{The coupling of an FPGA board to an opto-electronic reservoir was already reported in \cite{antonik2016online}, where the capacity to compute the output in real time was used to solve tasks that change in time. Note however that the FPGA design, i.e. the program implemented on the chip, differs greatly from the one used in the present work.}
\revv{Here we use the FPGA to feed the output of the reservoir back into itself, and thereby to generate periodic time signals and emulate chaotic systems. This is a completely different problem.}

Our experiments show that the system successfully solves two periodic time series generation tasks: frequency and random pattern generation, that have been previously investigated numerically \cite{caluwaerts2013locomotion,wyffels2014frequency,antonik2016towards2}. The first task allows to demonstrate different timescales within the neural network, and the second can be used as a memory metric. The photonic computer manages to generate both sine waves and random patterns with unlimited stability. Furthermore, we apply the RC to emulation of two chaotic attractors: Mackey-Glass \cite{mackey1977oscillation} and Lorenz \cite{lorenz1963deterministic} systems. In the literature, the performance on these tasks is quantified by computing the prediction horizon, i.e. the duration for which the RC can accurately follow a given trajectory on the chaotic attractor \cite{jaeger2004harnessing}. However, this method fails in the presence of a relatively high level of experimental noise, with a Signal-to-Noise Ratio (SNR) of roughly $40\units{dB}$, as will be discussed in section \ref{subsec:resnoise}. This noise was not problematic in previous experiments using the same opto-electronic reservoir \cite{paquot2012optoelectronic,antonik2016online}, but turns out to be intolerable for a system with output feedback. This raises the question of how to evaluate a system that emulates a known chaotic time series in the presence of noise. In this study, we introduce several new approaches, such as frequency spectrum comparison and randomness tests. \textcolor{rev1}{These approaches are based on well-known signal analysis techniques, but they are employed for the first time here for the evaluation of a chaotic signal generated by a reservoir computer.} Our results show that, although the RC struggles at following the target trajectory on the chaotic attractor, its output accurately reproduces the core characteristics of the target time series.

The paper is structured as follows. Sections \ref{sec:rc} and \ref{sec:tasks} introduce the basic principles of the reservoir computing and the time series generation tasks investigated in this work. The experimental setup, FPGA design and numerical simulations are outlined in section \ref{sec:exp}. All experimental and numerical results are presented and discussed in section \ref{sec:res}, and section \ref{sec:dis} concludes the paper.

\section{Basic principles of Reservoir Computing}
\label{sec:rc}

A typical reservoir computer contains a large number $N$ of internal variables $x_i(n)$ evolving in discrete time $n \in \mathbb{Z}$, as given by
\begin{equation}
  x_i(n+1) = f \left( \sum_{j=0}^{N-1} a_{ij} x_j(n) + b_i u(n) \right),
  \label{eq:rcevo}
\end{equation}
where $f$ is a nonlinear function, $u(n)$ is some external signal that is injected into the system, and $a_{ij}$ and $b_i$ are time-independent coefficients, drawn from some random distribution with zero mean, that determine the dynamics of the reservoir. The variances of these distributions are adjusted to obtain the best performances on the task considered.

The nonlinear function used here is $f = \sin (x)$, as in \cite{larger2012photonic, paquot2012optoelectronic}. To simplify the interconnection matrix $a_{ij}$, we exploit the ring topology, proposed in \cite{rodan2011minimum,appeltant2011information}, so that only the first neighbour nodes are connected. This architecture provides performances comparable to those obtained with complex interconnection matrices, as demonstrated numerically in \cite{lukovsevivcius2009survey,rodan2011minimum} and experimentally in \cite{appeltant2011information,larger2012photonic,paquot2012optoelectronic,duport2012all,brunner2012parallel}. Under these conditions Eq. \eqref{eq:rcevo} becomes
\begin{subequations}
  \begin{align}
    x_0(n+1) & = \sin \left(  \alpha x_{N-1}(n-1) + \beta M_0 u(n) \right),\label{eq:rcevo2_1} \\
    x_i(n+1) & = \sin \left(  \alpha x_{i-1}(n) + \beta M_i u(n) \right),\label{eq:rcevo2_2}
  \end{align}%
  \label{eq:rcevo2}%
\end{subequations}
\revs{with $i=1,\ldots,N-1$; $\alpha$ and $\beta$ parameters that are used to adjust the feedback and the input signals, respectively; and $M_i$ the input mask, drawn from a uniform distribution over the the interval $[-1, +1]$, as in \cite{rodan2011minimum, paquot2012optoelectronic, duport2012all}.}

The reservoir computer produces an output signal $y(n)$, given by a linear combination of the states of its internal variables
\begin{equation}
  y(n) = \sum_{i=0}^{N-1} w_i x_i (n),
  \label{eq:rcout}
\end{equation}
where $w_i$ are the readout weights, trained either offline (using standard linear regression methods, such as the ridge regression algorithm \cite{tikhonov1995numerical} used here), or online \cite{antonik2016online}, in order to minimise the Mean Square Error (MSE) between the output signal $y(n)$ and the target signal $d(n)$, given by
\begin{equation}
  \text{MSE} = \left\langle \left( y(n) - d(n) \right)^2 \right\rangle.
  \label{eq:mse}
\end{equation}

The introduction of the output feedback requires a minor change of notations. Since the RC can now receive two different signals as input, we shall \revs{denote} $I(n)$ the input signal, which can be either the external input signal $I(n) = u(n)$, or its own output, delayed by one timestep $I(n) = y(n-1)$.

\begin{figure}
  \centering
  \subfigure[Training stage]{\includegraphics[width=0.45\textwidth]{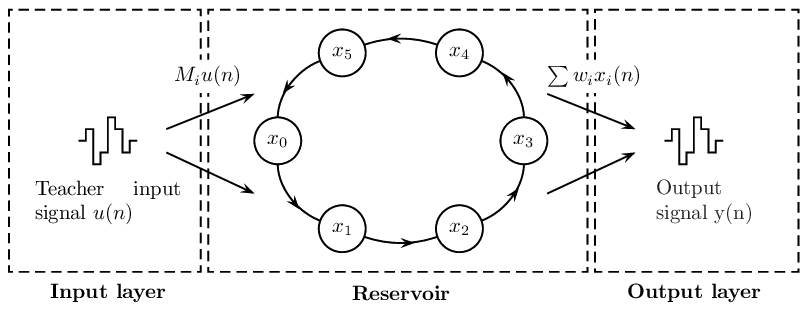}\label{subfig:rc}}
  \subfigure[Autonomous run]{\includegraphics[width=0.45\textwidth]{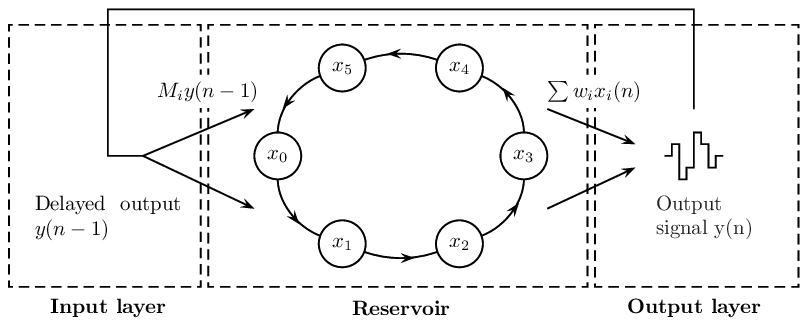}\label{subfig:rcfdb}}
  \caption{Schematic representation of the training stage \textbf{(a)} and the autonomous run \textbf{(b)} of our reservoir computer (here with $N=6$ nodes).
  \revs{During the training phase, the reservoir is driven by a teacher input signal $u(n)$, and the readout weights $w_i$ are optimised for the output to be as close as possible to $u(n+1)$}.
  During the autonomous run, the teacher signal is switched off and the reservoir is driven by its own output signal. The readout weights $w_i$ are kept constant and the performance of the system is measured in terms of how long or how well it can generate the desired output.
}
  \label{fig:rc}
\end{figure}

The reservoir computer is operated in two stages, depicted in Fig. \ref{fig:rc}: a training phase and an autonomous run. 
During the training phase, the reservoir computer is driven by a time-multiplexed teacher signal $I(n) = u(n)$, and the resulting states of the internal variables $x_i(n)$ are recorded. The teacher signal depends on the task under investigation (which will be introduced in section \ref{sec:tasks}). The system is trained to predict the next value of the teacher time series from the current one, that is, the readout weights $w_i$ are optimised so as to get as close as possible to $y(n) = u(n+1)$.
Then, the reservoir input is switched from the teacher sequence to the reservoir output signal $I(n) = y(n-1)$, and the system is left running autonomously. The reservoir output $y(n)$ is used to evaluate the performance of the experiment.

\section{Time series generation tasks}
\label{sec:tasks}

Feeding the output back into the reservoir allows the computer to autonomously (i.e. without any external input) generate time series. We tested the capacity of the experiment to generate both periodic and chaotic signals, with two tasks in each category.

\subsection{Frequency generation}
\label{subsec:taskfreq}
Frequency generation is the simplest time series generation task considered here. The system is trained to generate a sine wave given by
\begin{equation}
  u(n) = \sin\left( \nu n \right),
  \label{eq:freq}
\end{equation}
where $\nu$ is a real-valued relative frequency. The physical frequency $f$ of the sine wave depends on the experimental roundtrip time $T$ (see section \ref{sec:exp}) as follows
\begin{equation}
  f = \frac{\nu}{2\pi T}.
  \label{eq:freq2}
\end{equation}
This task allows to measure the bandwidth of the system and investigate different timescales within the neural network.

\subsection{Random pattern generation}
\label{subsec:taskpat}
Random pattern generation is a natural step forward from the frequency generation task to a more complex problem -- instead of a regularly-shaped continuous function, the system is trained to generate an arbitrarily-shaped discontinuous function (that remains periodic, though). Specifically, a pattern is a short sequence of $L$ randomly chosen real numbers (here within the interval $\left[-0.5, 0.5 \right]$) that is repeated periodically to form an infinite time series \cite{antonik2016towards}.
Similarly to the physical frequency in section \ref{subsec:taskfreq}, the physical period of the pattern is given by $\tau_\text{pattern} = L \cdot T$.
The aim is to obtain a stable pattern generator, that reproduces precisely the pattern and doesn't deviate to another periodic behaviour. To evaluate the performance of the generator, we compute the MSE between the reservoir output signal and the target pattern signal during the training phase and the autonomous run, with a maximal threshold set to $10^{-3}$. This value is somewhat arbitrary, and one could have picked a different threshold. As will be illustrated in Figs. \ref{fig:reslarge} and \ref{fig:lzpred} in Sec. \ref{sec:res}, the $10^{-3}$ level corresponds to the point where the RC strongly deviates from the starting trajectory on the chaotic attractor. 
For consistency, we have used this threshold in all our experiments, for all tasks. If the error doesn't grow above the threshold during the autonomous run, the system is considered to accurately generate the target pattern. We also tested the long-term stability on several patterns by running the system for several hours, as will be described in section \ref{sec:res}.

\subsection{Mackey-Glass chaotic series prediction}
\label{subsec:taskmg}
The Mackey-Glass delay differential equation
\begin{equation}
  \frac{dx}{dt} = \beta \frac{x(t-\tau)}{1+x^n(t-\tau)} - \gamma x
  \label{eq:mg}
\end{equation}
with $\tau$, $\gamma$, $\beta$, $n > 0$ was introduced to illustrate the appearance of complex dynamics in physiological control systems \cite{mackey1977oscillation}. To obtain chaotic dynamics, we set the parameters as in \cite{jaeger2004harnessing}: $\beta = 0.2$, $\gamma = 0.1$, $\tau = 17$ and $n=10$. 
\textcolor{rev1}{With these characteristics, the Kaplan-Yorke dimension of the chaotic attractor is $2.1$ \cite{farmer1982chaotic}.}

The equation was solved using the Runge-Kutta 4 method with a stepsize of $1.0$. To avoid unnecessary computations and save time, both in simulations and experiments, we pre-generated a sequence of $10^6$ samples that we used for all numerical and experimental investigations.

During autonomous run, without the correct teacher signal, the system slowly deviates from the desired trajectory. The MSE is used to evaluate both the training phase and the autonomous run. We then compute the number of correct prediction steps, i.e. steps for which the MSE stays below the  $10^{-3}$ threshold (see section \ref{subsec:taskpat}), during the autonomous run and use this figure to evaluate the performance of the system.

\subsection{Lorenz chaotic series prediction}
\label{subsec:tasklz}
The Lorenz equations, a system of three ordinary differential equations 
\begin{subequations}
  \begin{align}
    \frac{dx}{dt} & = \sigma \left( y - x \right), \\
    \frac{dy}{dt} & = -xz + rx - y, \\
    \frac{dz}{dt} & = xy - bz,
  \end{align}
\end{subequations}
with $\sigma, r, b > 0$, was introduced as a simple model for atmospheric convection \cite{lorenz1963deterministic}. The system exhibits chaotic behaviour for $\sigma=10, b=8/3$ and $r=28$ \cite{hirsch2003differential}, that we used in this study. This yields a chaotic attractor with the highest Lyapunov exponent of $\lambda=0.906$ \cite{jaeger2004harnessing}. The system was solved using Matlab's \verb+ode45+ solver and a stepsize of $0.02$, as in \cite{jaeger2004harnessing}. We used all computed points, meaning that one timestep of the reservoir computer corresponds to a step of $0.02$ in the Lorenz time scale. 
\revs{To avoid unnecessary computations and save time we pre-generated a sequence of $10^5$ samples that we used for all numerical and experimental investigations.}
Following \cite{jaeger2004harnessing}, we used the $x$-coordinate trajectory for training and testing, that we scaled by a factor of $0.01$.

\section{Experimental setup}
\label{sec:exp}

\begin{figure*}
  \centering
  \includegraphics[width=0.7\textwidth]{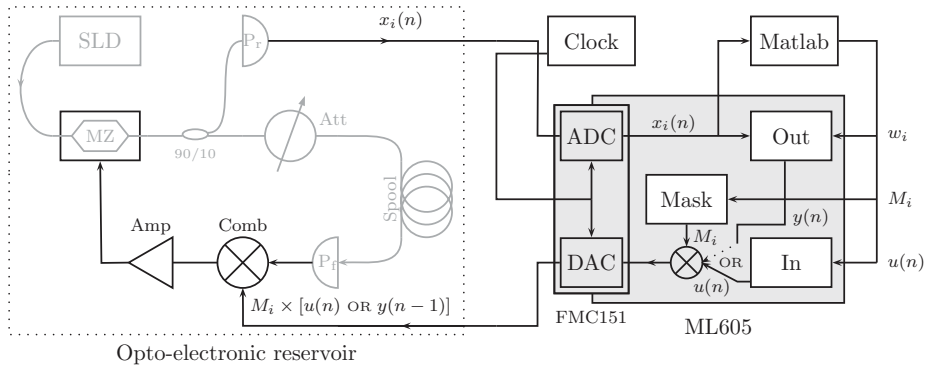}
  \caption{Schematic representation of the experimental setup. Optical and electronic components of the photonic reservoir are shown in grey and black, respectively. It contains an incoherent light source (SLD), a Mach-Zehnder intensity modulator (MZ), a $90/10$ beam splitter, an optical attenuator (Att), a fibre spool (Spool), two photodiodes ($\text{P}_\text{r}$ and $\text{P}_\text{f}$), a resistive combiner (Comb) and an amplifier (Amp). The FPGA board implements the readout layer and computes the output signal $y(n)$ in real time. It also generates the analogue input signal $I(n)$ and acquires the reservoir states $x_i(n)$.
  The computer, running Matlab, controls the devices, performs the offline training and uploads all the data ($u(n)$, $w_i$ and $M_i$) on the FPGA.
  }
  \label{fig:exp}
\end{figure*}

Our experimental setup, schematised in Fig. \ref{fig:exp}, consists of two main components: the opto-electronic reservoir and the FPGA board.

\subsection{Opto-electronic reservoir}
\label{subsec:oeres}

The opto-electronic reservoir is based on previously published works \cite{paquot2012optoelectronic,larger2012photonic}. The reservoir states are encoded into the intensity of an incoherent light signal, produced by a superluminiscent diode (Thorlabs SLD1550P-A40). The Mach-Zehnder (MZ) intensity modulator (EOSPACE AX-2X2-0MSS-12) implements the nonlinear function, its operating point is adjusted by applying a bias voltage, produced by a Hameg HMP4040 power supply.  A fraction (10\%) of the signal is extracted from the loop and sent to the readout photodiode (TTI TIA-525I) and the resulting voltage signal is sent to the FPGA. An optical attenuator (Agilent 81571A) is used to set the feedback gain $\alpha$ of the system (see Eqs. \eqref{eq:rcevo2_1} and \eqref{eq:rcevo2_2}). 
The resistive combiner sums the electrical feedback signal, produced by the feedback photodiode (TTI TIA-525I), with the input signal from the FPGA to drive the MZ modulator, with an additional amplification stage of $+27 \units{dB}$ (ZHL-32A+ coaxial amplifier) to span the entire $V_\pi$ interval of the modulator. 

As the neurons are time-multiplexed, the maximal reservoir size depends on the delay from the fibre spool (Spool) and the sampling frequency of the Analogue-to-Digital converter (ADC). While increasing the latter involves relatively high costs, one can lengthen the delay line fairly easily. In this work, we used two spools of single mode fibre of lengths 1.6 km and 10 km, approximately. The first produced a delay of $7.93\units{\textmu s}$ and allowed to fit $N=100$ neurons into the reservoir. The second spool was used to increase the delay up to $49.2\units{\textmu s}$ and the reservoir size up to $N=600$. In both cases, the reservoir states were sampled at approximately $200\units{MHz}$ (the precise frequency, given in section \ref{subsec:fpga}, depends on the delay loop) and each state was averaged over 16 samples in order to decrease the noise and remove the transients induced by the finite bandwidth of the Digital-to-Analogue converter (DAC).

The experiment is operated as follows.
First, the input mask $M_i$ and the teacher signal $u(n)$, generated in Matlab, are uploaded on the FPGA board, which then generates the masked input signal $M_i \times u(n)$, sent to the reservoir via the DAC. The resulting reservoir states $x_i(n)$ are sampled and averaged by the FPGA, and then sent to the computer in real time. That is, the design allows to capture the reservoir states for any desired time interval. After training of the reservoir,
the optimal readout weights $w_i$ are uploaded on the board.
Because of the relatively long delay needed for the offline training, the reservoir needs to be reinitialised in order to restore the desired dynamics of the internal states prior to running it autonomously. For this reason, we drive the system with an initialisation sequence of $128$ timesteps (as illustrated in Fig. \ref{fig:expat}), before coupling the output signal with the input and letting the reservoir computer run autonomously. 
In this stage, the FPGA computes the output signal $y(n)$ in real time, then creates a masked version $M_i \times y(n)$ and sends it to the reservoir via the DAC.

As the neurons are processed sequentially, due to propagation delay between the intensity modulator (MZ) and the ADC, the output signal $y(n)$ can only be computed in time to update the 24-th neuron $x_{23} (n+1)$. For this reason, we set the first 23 elements of the input mask $M_i$ to zero. That way, all neurons contribute to solving the task, but the first 23 do not ``see'' the input signal $I(n)$. \revv{Note that this reflects an aspect that is inherent to any experimental implementation of time-multiplexed reservoir computing with feedback. In principle, the output $y(n)$ has to be computed after the acquisition of the last neuron $x_{N-1} (n)$ at timestep $n$, but before the first neuron $x_0 (n+1)$ of the following timestep. However, in time-multiplexed RC implementations, these states are consecutive, and the experiment cannot be paused to let $y(n)$ be computed.
There may therefore be a delay (whose duration depends on the hardware used) before $y(n)$ is computed and can be fed back into the reservoir. In the present experiment, this delay is approximately $115\units{ns}$, which corresponds to 23 neuron durations.}
As we will see below this has an impact on system performance.

\subsection{FPGA board}
\label{subsec:fpga}

In this work we use a Xilinx ML605 evaluation board, powered by a Virtex 6 XC6VLX240T FPGA chip. The board is paired with a 4DSP FMC151 daughter card, containing one two-channel ADC and one two-channel DAC. The ADC's maximum sampling frequency is $250 \units{MHz}$ with 14-bit resolution, while the DAC can sample at up to $800 \units{MHz}$ with 16-bit precision. 

The FPGA design is written in standard IEEE 1076-1993 VHDL language \cite{ieeevhdl, pedroni2004circuit} and compiled with Xilinx ISE Design Suite 14.7, provided with the board. We also used Xilinx ChipScope Pro Analyser to monitor signals on the board, mostly for debugging and testing.

\begin{figure*}
  \centering
  \includegraphics[width=0.70\textwidth]{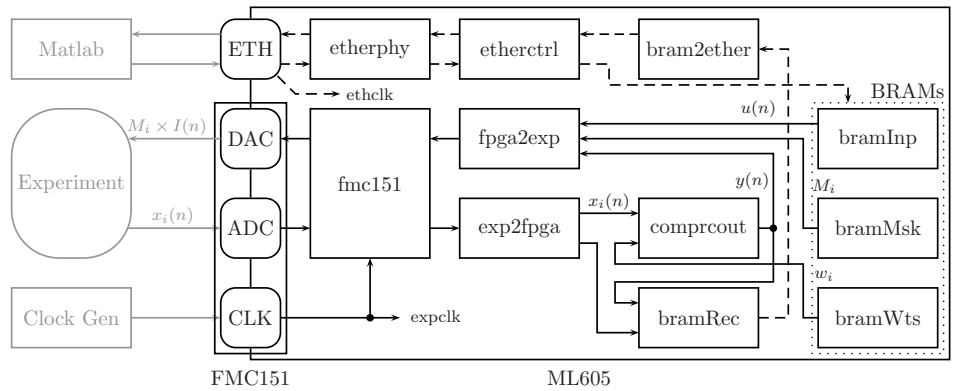}
  \cprotect\caption{Simplified schematics of the FPGA design. Modules (entities) are represented by rectangular boxes, onboard electronic components are shown with rounded rectangles. External hardware, such as the computer, running Matlab, the opto-electronic reservoir and the external clock generator are shown in grey. The design is driven by two clocks: the experimental clock \verb+expclk+ and the Ethernet clock \verb+ethclk+. Signals from these two clock domains are drawn in solid and dashed lines, respectively.}
  \label{fig:fpga}
\end{figure*}

The simplified schematics of the design is depicted in Fig. \ref{fig:fpga}. Rectangular boxes represent modules (entities), and rounded rectangles stand for electronic components on the ML605 board, namely the FMC151 daughtercard and the onboard Marvell Alaska PHY device (88E1111) for Ethernet communications (ETH).

The operation of the FPGA is controlled from the computer, running Matlab, via a simple custom protocol through a Gbit Ethernet connection. Data and various commands, such as memory read/write, or state change, are encapsulated into standard UDP packages. The \verb+etherphy+ module interfaces the FPGA design with the Marvell Ethernet PHY chip, and the \verb+etherctrl+ module receives the UDP packets (frames) and decodes the commands and the data. It also creates the frames for sending data from FPGA to the computer.

Blocks of Random-Access Memory (BRAM) are used to store data, such as teacher inputs $u(n)$, input masks $M_i$ and readout weights $w_i$, that are generated on the computer and uploaded on the FPGA. Each type of data is assigned a specific module, since they vary in size (e.g. 600 values for the input mask and up to 3000 for the teacher signal) and resolution, as will be explained below. The \verb+bramRec+ is a buffer-like module, designed to transfer the signal recorded from the experiment directly to the computer through Ethernet, without permanently storing it in memory. It consists of two blocks of RAM of 2048 bytes each, that are used as follows: while the recorded signal is written into the first block, \verb+bram2ether+ reads the contents of the second, that is then encapsulated into four UDP frames sent to the computer. When the first block is full, the blocks are switched and the process continues.

The FMC151 daughtercard is interfaced with the rest of the design through the \verb+fmc151+ module, that outputs two 14-bit signals from the ADCs and receives two 16-bit signals from the DACs. The FMC151 card is also used to deliver a clock signal from an external clock generator, that produces a high-precision signal, allowing to synchronise the FPGA with the delay loop of the experimental setup. This clock signal was generated by the Hewlett Packard 8648A signal generator. In our experiments with two delay loops (see section \ref{subsec:oeres}), we fine-tuned the clock frequency so as to fit 16 samples per neuron into the roundtrip time $T$. Specifically, we sampled the $N=600$ reservoir states at $195.4472\units{MHz}$ with a large fibre spool, and at $203.7831\units{MHz}$ with a small spool and $N=100$.

The \verb+fpga2exp+ module controls the signal sent to the opto-electronic reservoir through the DAC. At the training phase, it generates the masked input signal $M_i \times u(n)$ by multiplying the inputs $u(n)$ by the mask $M_i$, both being read from the BRAMs. During the autonomous run, it receives the reservoir output signal $y(n)$, computed by the \verb+comprcout+ module, masks it and transfers to the DAC. 

The neuron states $x_i(n)$ from the photonic reservoir are sampled and averaged by the \verb+exp2fpga+ module. At the training phase, these are buffered in \verb+bramRec+, then processed by the \verb+bram2ether+ module and sent to the computer. During the autonomous run, the reservoir states are used by the \verb+comprcout+, together with the readout weights $w_i$, read from the \verb+bramWts+ memory, to compute the reservoir output $y(n)$. It is then injected back into the reservoir through the \verb+fpga2exp+, and also transferred to the computer through the \verb+bramRec+ and \verb+bram2ether+ modules.

The design is driven by two clocks: the experimental clock \verb+expclk+ (around $200\units{MHz}$, depending on the loop delay $T$) that operates data acquisition and generation modules and allows to synchronise the FPGA with the experiment, and the $125\units{MHz}$ Ethernet clock \verb+ethclk+. Both clocks have to be managed properly within the design, as several signals, such as inputs or weights, coexist in both clock domains. That is, data to BRAMs comes from the Ethernet modules, and is thus driven by the \verb+ethclk+ clock. On the other hand, this same data is used by the \verb+fpga2exp+ module, and has to appear in the \verb+expclk+ clock domain. To this end, we exploit the dual-port capability of Xilinx block RAMs. That is, data is written into memory blocks through port A at clock \verb+ethclk+ and read from port B at clock \verb+expclk+ (and vice versa for the \verb+bramRec+). This allows for smooth transition of data between clock domains. The two clock domains are depicted in Fig. \ref{fig:fpga} as follows: signals running at \verb+expclk+ are shown in solid lines, and those clocking at \verb+ethclk+ are drawn with dashed lines.

The arithmetic operations computed by the FPGA are performed on real numbers. However, the chip is a logic device, designed to operate bits. The performance of the design thus highly depends on the bit-representation of real numbers, i.e. the precision. The main constraint comes from the ADC and DAC, limited to 14 and 16 bits, respectively. Numerical simulations, reported in \cite{antonik2016towards}, show that such precision is sufficient for all tasks studied in this work. It was also shown in \cite{antonik2016towards} that the precision of the readout weights $w_i$ has a significant impact on the performance of the system. For this reason we designed the experiment for optimal utilisation of the resolution available. The reservoir states were tuned to lie within a $]-1,+1[$ interval. They are thus represented as 16-bit integers, with 1 bit for the sign and 15 bits for the decimal part. Another limitation comes from DSP48E slices, used to multiply the states $x_i(n)$ by the readout weights $w_i$. These blocks are designed to multiply a 25-bit integer by a 18-bit integer. To meet these requirements, we also keep the readout weights $w_i$ within the $]-1,1[$ interval and represent them as 25-bit integers, with 1 sign bit and 24 decimal bits. To ensure that $w_i \in ]-1,1[$, we amplify the reservoir states digitally inside the FPGA. That is, the $x_i(n)$ are multiplied by 8 after acquisition, prior to computing the output signal $y(n)$.

\subsection{Numerical simulations}

In addition to the physical experiments, we investigated the proposed setup in numerical simulations, to have a point of comparison and identify possible malfunctions. To this end, we developed three models that simulate the experiment to different degrees of accuracy. Our custom Matlab scripts are based on \cite{paquot2012optoelectronic,antonik2016towards}.

\begin{description}
  \item[Idealised model] 
    It incorporates the core characteristics of our reservoir computer, i.e. the ring-like architecture, the sine nonlinearity and the linear readout layer (as described by equations \ref{eq:rcevo2} and \ref{eq:rcout}), disregarding all experimental considerations. We use this model to define the maximal performance achievable in each configuration.
  \item[Noiseless experimental model]
    This model emulates the most influential features of the experimental setup, such as the high-pass filter of the amplifier, the finite resolution of the ADC and DAC, and precise input and feedback gains. 
    This model allows to cross-check the experimental results and to easily identify the problematic points.
  \item[Noisy experimental model]
    Contrary to the noiseless numerical models introduced above, our experimental implementation is noisy, which, as will be explained below, has a significant impact on performance. 
    In order to compare our experimental results to a more realistic model, we estimated the level of noise present in the experimental system (see section \ref{subsec:resnoise}), and incorporated this noise into the noisy version of the experimental model.
\end{description}

\section{Results}
\label{sec:res}

In this section we present the experimental results, compare them to numerical simulations and discuss the performance of the reservoir computer on each task introduced in section \ref{sec:tasks}.

The two periodic signal generation tasks were solved using a small reservoir with $N=100$ and a fibre spool of approximately $1.6\units{km}$. The chaotic signal generation tasks, being more complex, required a large reservoir of $N=600$ for decent results, that we fit in a delay line of roughly $10\units{km}$.

\subsection{Noisy reservoir}
\label{subsec:resnoise}

For most tasks studied here, we found the experimental noise to be the major source of performance degradation in comparison to numerical investigations. 
In fact, previously reported simulations \cite{antonik2016towards} considered an ideal noiseless reservoir, while our experiment is noisy. 
This noise is generated by the active components of the setup: the amplifier, which has a relatively high gain and is therefore very sensitive to small parasitic signals on the input, the DAC and the photodiodes. 
In-depth experimental investigations (not presented in this paper) have shown that, in fact, each component contributes more or less equally to the overall noise level. Thus, it can not be reduced by replacing one ``faulty'' component. Neither can it be averaged out, as the output value has to be computed at each timestep. This noise was found to have a significant impact on the results, as will be shown in the following sections. For this reason we estimated the level of noise present in the experimental system and incorporated it to the numerical models. This allows us to ``switch off'' the noise in simulations, which is impossible experimentally.

\begin{figure}
  \centering
  \includegraphics[width=0.47\textwidth]{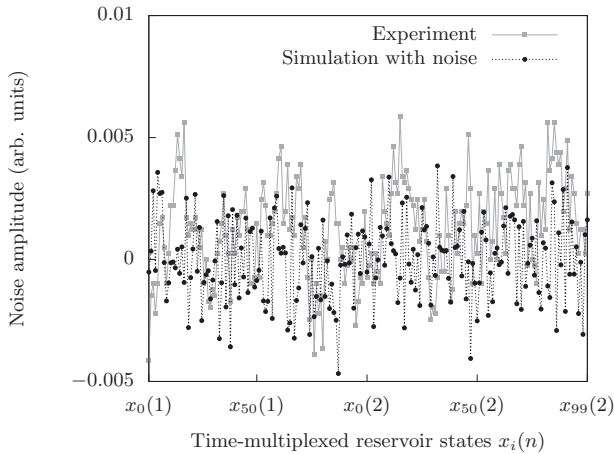}
  \caption{
    Illustration of the noise inside the experimental reservoir. Experimental (solid grey line) and numerical (dotted black curve) reservoir states $x_i (n)$ are shown in the case when the input signal is null $I(n)=0$, scaled so that in normal experimental conditions (non-zero input) they would lie in a $[-1,1]$ interval. Although the input signal is null $I(n)=0$, the actual neurons are non-zero because of noise. Numerical noise was generated with a Gaussian random distribution with standard deviation of $1\e{-3}$ so that to reproduce the noise level of the experiment.
  }
  \label{fig:noise2}
\end{figure}

Fig. \ref{fig:noise2} shows numerical and experimental reservoir states of a 100-neuron reservoir, as received by the readout photodiode. That is, the curves depict the time-multiplexed neurons: each point represents a reservoir state $x_{0\ldots99}(n)$ at times $n=1$ and $n=2$. The system does not receive any input signal $I(n)=0$. The experimental signal is plotted with a solid grey line. We use it to compute the experimental noise level by taking the standard deviation of the signal, which gives $2\e{-3}$. We then employed this noise level in the noisy experimental model to compare experimental results to numerical simulations. The dotted black curve in Fig. \ref{fig:noise2} shows the response of the noisy experimental model, with the same amount of Gaussian noise (standard deviation of $2.0\e{-3}$) as in the experiment. \textcolor{rev1}{The choice of a Gaussian noise distribution was validated by experimental measurements.}

The level of the experimental noise can also be characterised by the Signal-to-Noise Ratio (SNR), defined as \cite{horowitz1980art}
\[
  \text{SNR} = 10 \log_{10} \left( \frac{ \text{RMS}_\text{signal}^2 }{ \text{RMS}_\text{noise}^2 } \right),  
\]
where RMS is the Root Mean Square value, given by
\[
  \text{RMS}(x_i) = \sqrt{ \frac{1}{N} \sum_{i=1}^{N} x_i^2 }.
\]
We measured $\text{RMS}_\text{signal}=0.2468$ and $\text{RMS}_\text{noise}=0.0023$, so the SNR is equal to approximately $40\units{dB}$ in this case.
Note that this figure is given as an indicator of order of magnitude only \revs{as the} RMS of the reservoir states depends on the gain parameters ($\alpha$ and $\beta$, see section \ref{sec:rc}) and varies from one experiment to another.

\subsection{Frequency generation}

We found the frequency generation task to be the only one not affected by noise. That is, our experimental results matched accurately the numerical predictions reported in \cite{antonik2016towards2}. Concretely, we expected a bandwidth of $\nu \in \left[ 0.06, \pi \right]$ with a 100-neuron reservoir.
The upper limit is a signal oscillating between $-1$ and $1$ and is given by half of the sampling rate of the system (the Nyquist frequency). The lower limit is caused by the memory limitation of the reservoir. In fact, low-frequency oscillations correspond to longer periods, and the neural network can no longer ``remember'' a sufficiently long segment of the sine wave so as to keep generating a sinusoidal output.
These numerical results are confirmed experimentally here.

\begin{figure}
  \centering
    \includegraphics[width=0.45\textwidth]{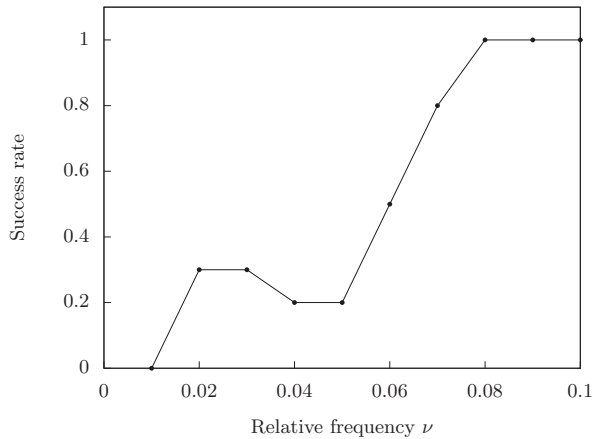}
    \caption{Determination of the lower limit of the reservoir computer bandwidth. Frequencies above $0.08$ are generated very well with any of the 10 random input mask, and are therefore not shown on the plot. Frequencies below $0.05$ fail with most input masks. We thus consider $0.08$ as the lower limit of the bandwidth, but keep in mind that frequencies as low as $0.02$ could also be generated, but only with a carefully picked input mask.
}
  \label{fig:bw}
\end{figure}

We tested our setup on frequencies \revs{$\nu$} ranging from $0.01$ to $\pi$, and found that frequencies within $[0.1, \pi]$ are generate accurately with any random input mask. Lower frequencies between $0.01$ and $0.1$, however, were produced properly with some random masks, but not all. For this reason, we investigated the $[0.01, 0.1]$ interval more precisely, since this is where the lower limit of the bandwidth lies. For each frequency, we ran the experiment 10 times for 10k timesteps with different random input masks and counted the number of times the reservoir produced a sine wave with the desired frequency ($\text{MSE} < 10^{-3}$, see section \ref{subsec:taskpat}) and amplitude of $1$. The results are shown in Fig. \ref{fig:bw}. Frequencies below 0.05 are not generated correctly with most input masks. At $\nu=0.7$ the output is correct most of the times, and for $\nu=0.08$ and above the output sine wave is correct with any input mask. The bandwidth of this experimental RC is thus $\nu \in \left[ 0.08, \pi \right]$. 
Given the roundtrip time $T=7.93\units{\textmu s}$, this results in a physical bandwidth of $1.5$ -- $63\units{kHz}$.
Note that frequencies within this interval can be generated with any random input mask $M_i$. Lower frequencies, down to $0.02$, could also be generated, but only with a suitable input mask.

\begin{figure}
  \centering
  \includegraphics[width=0.45\textwidth]{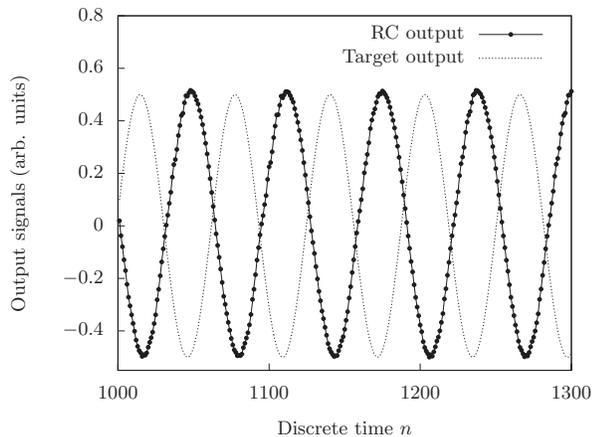}
  \caption{Example of an autonomous run output signal for frequency generation task with $\nu=0.1$. The experiment continues beyond the range of the figure.}
  \label{fig:exfreq}
\end{figure}

Fig. \ref{fig:exfreq} shows an example of the output signal during the autonomous run. The system was trained for $1000$ timesteps to generate a frequency of $\nu = 0.1$, and successfully accomplished this task with a MSE of $5.6\e{-9}$.

\revs{The above results were obtained by scanning the input gain $\beta$ and the feedback gain $\alpha$ to obtain the best results. It was found that $\beta$ has little impact on the system performance so long as it is chosen in the interval $\beta\in[0.02,0.5]$, while the feedback gain $\alpha$, on the contrary, has to lie within a narrow interval, approximately between $0.9$ and $1.0$, otherwise the reservoir yields very poor results.}

\subsection{Random pattern generation}

\begin{figure}
  \centering
  \includegraphics[width=0.45\textwidth]{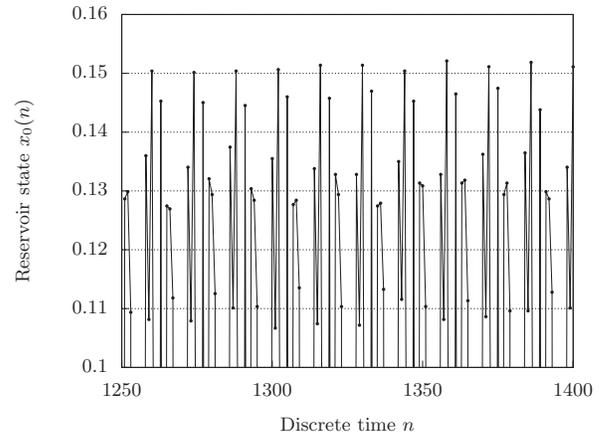}
  \caption{Example of behaviour of one neuron in a noisy experimental reservoir. For clarity, the range of the Y axis is limited to the area of interest. Because of noise, despite a periodic input signal $u(n)$, the reservoir state takes similar, but not identical values.
  }
  \label{fig:resnoisynrn}
\end{figure}

The random pattern generation task is more complex than frequency generation and is slightly affected by the experimental noise. The goal of this task is two-fold: ``remember'' a pattern of a given length $L$ and be able to reproduce it for an unlimited duration. We have shown numerically that a noiseless 51-neuron reservoir is capable of generating patterns up to 51-element long \cite{antonik2016towards}. This is a logical result, as, intuitively, each neuron of the system is expected to ``memorise'' one value of the pattern.
Simulations of a noisy 100-neuron reservoir, similar to the experimental setup, show that the maximum pattern length is reduced down to $L=13$. This means that noise significantly reduces the effective memory of the system. In fact, the noisy neural network has to take into account the slight deviations of the output from the target pattern so as to be able to follow the pattern disregarding these imperfections. Fig. \ref{fig:resnoisynrn} illustrates this issue. Periodic oscillations of one neuron of the reservoir are shown, with intended focus on the upper values and an adequate magnification so as to see the small variations. It shows that the neuron oscillates between similar, but not identical values. This makes the generation task much more complex, and requires more memory, hence the maximal pattern length is shorter. 

\begin{figure*}
  \centering
  \includegraphics[width=0.90\textwidth]{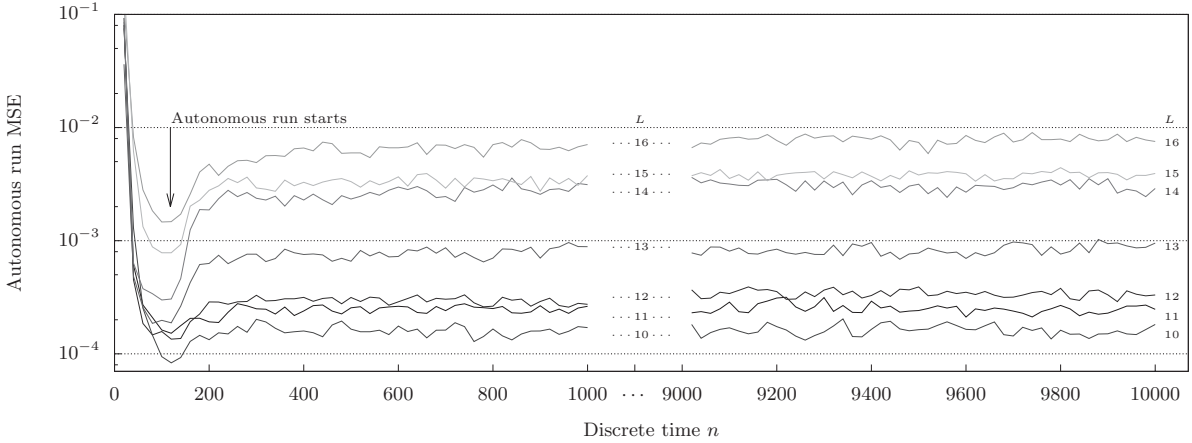}
    \caption{%
     Evolution of MSE during experimental autonomous generation of periodic random patterns of lengths $L=10, \ldots, 16$ as function of discrete time $n$. The autonomous run starts at $n=128$, as indicated by the arrow. Patterns shorter than 13 are reproduced with low $\text{MSE}<10^{-3}$, while patterns longer than 14 are not generated correctly with $\text{MSE}>10^{-3}$. In the latter cases, the reservoir dynamics remains stable and periodic, but the output only remotely resembles the target pattern.}
  \label{fig:errevo}
\end{figure*}

We obtained similar results in the experiments. Fig. \ref{fig:errevo} shows the evolution of the MSE measured during the first 1k timesteps of 10k-timestep autonomous runs with different pattern lengths. Plotted curves are averaged over 100 runs of the experiment, with 5 random input masks and 20 random patterns for each length. The initial minimum (at $n=128$) corresponds to the initialisation of the reservoir (see section \ref{subsec:oeres}), then the output is coupled back and the system runs autonomously. Patterns with $L=12$ or less are generated very well and the error stays low. Patterns of length 13 show an increase in MSE, but they are still generated reasonably well. For longer patterns, the system deviates to a different periodic behaviour, and the error grows above $10^{-3}$.

\begin{figure}
  \centering
  \includegraphics[width=0.45\textwidth]{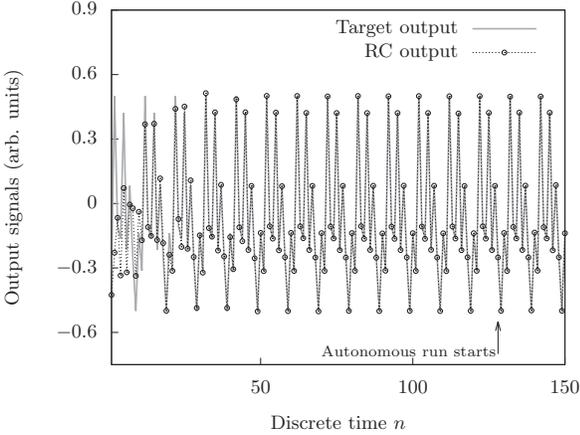}
  \caption{Example of an output signal for random pattern generation task, with a pattern of length 10. The reservoir is first driven by the desired signal for $128$ timesteps (see section \ref{subsec:fpga}), and then the input is connected to the output. Note that in this example the reservoir output requires about 50 timesteps to match the driver signal. The autonomous run continues beyond the scope of the figure.}
  \label{fig:expat}
\end{figure}

\begin{figure}
  \centering
  \includegraphics[width=0.45\textwidth]{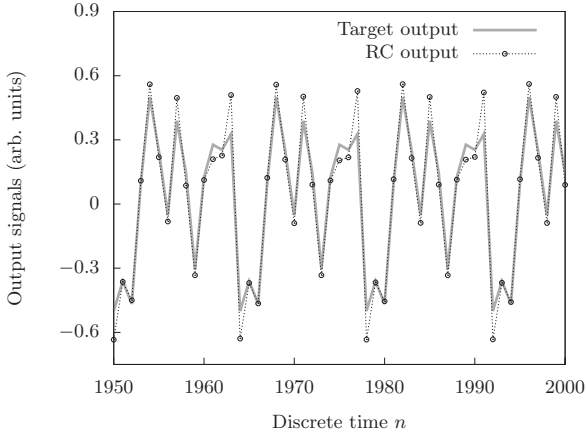}
  \caption{Example of an autonomous run output after 1950 timesteps, with a pattern of length $L=14$. The RC outputs a periodic signal that clearly does not match the target pattern ($\text{MSE}=5.2\e{-3}$).}
  \label{fig:expat2}
\end{figure}

Fig. \ref{fig:expat} shows an example of the output signal during the autonomous run. The system was trained for $1000$ timesteps to generate a pattern of length 10. The reservoir computer successfully learned the desired pattern and the output perfectly matches the target signal. Fig. \ref{fig:expat2} illustrates a case with a longer patter ($L=14$), that could not be learned by the system. As can be seen from the plot, the RC captured the general shape of the pattern, but can not accurately generate individual points. The MSE of this run is $5.2\e{-3}$, which is above the acceptable $10^{-3}$ threshold.

We also tested the stability of the generator by running it for several hours ($\sim10^9$ timesteps) with random patterns of lengths 10, 11 and 12. The output signal was visualised on a scope and remained stable and accurate through the whole test.

\revs{The above results were obtained by scanning the input gain $\beta$ and the feedback gain $\alpha$ to obtain the best results. As for frequency generation, it was found that $\beta$ has little impact on the system performance so long as it is chosen in the interval $\beta\in[0.1,1]$, while the feedback gain $\alpha$, on the contrary, has to lie within a narrow interval, approximately between $0.9$ and $1.0$.}

\subsection{Mackey-Glass series prediction}

Chaotic time series generation tasks were the most affected by the experimental noise. This is not surprising, since chaotic systems are, by definition, very sensitive to noise. Reservoir computing was first applied to this class of tasks in \cite{jaeger2004harnessing}. In their numerical work, the authors investigated the capacity of the computer to follow a given trajectory in the phase space of the chaotic attractor. We also tried this approach, but since our experimental system performs as a ``noisy'' emulator of the chaotic attractor, its trajectory deviates very quickly from the target one, especially with a SNR as low as $40\units{dB}$ (see section \ref{subsec:resnoise}). For this reason, we considered different approaches to evaluate the performance of the system, as will be described below.

The system was trained over 1500 input samples and was running autonomously for 600 timesteps. In particular, we prepared 2100 steps of the Mackey-Glass series for each run of the experiment and used the first 1500 as a teacher signal $u(n)$ to train the system and the last 600 both as an initialisation sequence (see section \ref{subsec:fpga}) and as a target signal $d(n)$ to compute the MSE of the output signal $y(n)$. These 2100 samples were taken from several starting points $t$ (see equation \eqref{eq:mg}) in order to test the reservoir computer on different instances of the Mackey-Glass series.
We scanned the input gain and the feedback attenuation ($\beta$ and $\alpha$ in equations \eqref{eq:rcevo}) to find optimal dynamics of the opto-electronic reservoir for this task. We used $\beta \in [0.1, 0.3] $ and tuned the optical attenuator in the range $[4.25, 5.25] \units{dB}$, which corresponds approximately to $\alpha \in [0.85, 0.95]$, with slightly different values for different instances of the Mackey-Glass series. 

\begin{figure}
  \centering
  \includegraphics[width=0.45\textwidth]{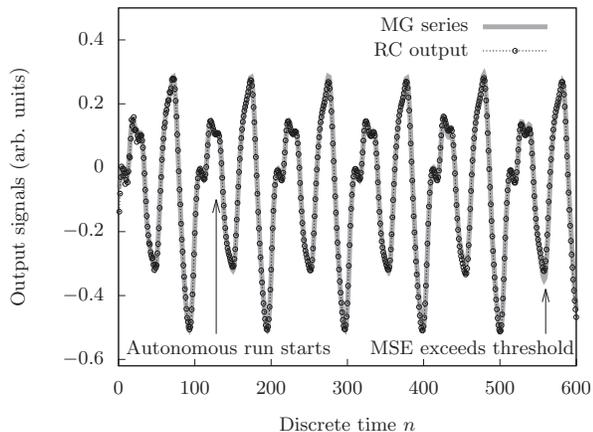}
  \caption{Example of reservoir computer output signal $y(n)$ (dotted black line) during autonomous run on the Mackey-Glass task. The system was driven by the target signal (solid grey line) for 128 timesteps and then left running autonomously, with $y(n)$ coupled to the input $I(n)$ (see Main text). The MSE threshold was set to $10^{-3}$. The photonic reservoir computer with $N=600$ was able to generate up to 435 correct values.}
  \label{fig:reslarge}
\end{figure}

\begin{figure}
  \centering
  \includegraphics[width=0.45\textwidth]{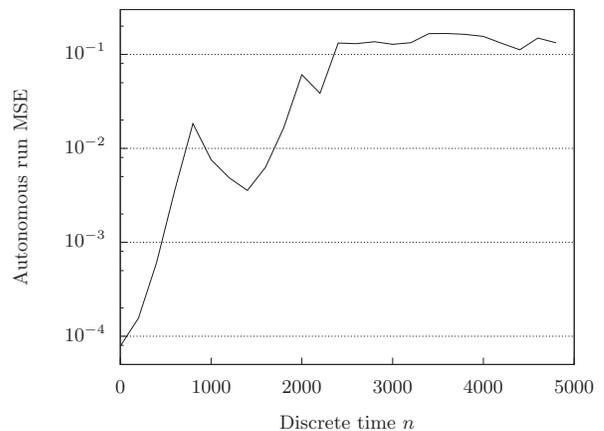}
  \caption{\textcolor{rev1}{
    Evolution of MSE during experimental autonomous generation of the Mackey-Glass chaotic time series \revs{(same run as in Fig. (\ref{fig:reslarge})}. The error curve, averaged over 200 timesteps, crosses the $10^{-3}$ threshold approximately between $n=500$ and $n=600$.
   }}
  \label{fig:mgerrevo}
\end{figure}

\begin{figure}
  \centering
  \includegraphics[width=0.45\textwidth]{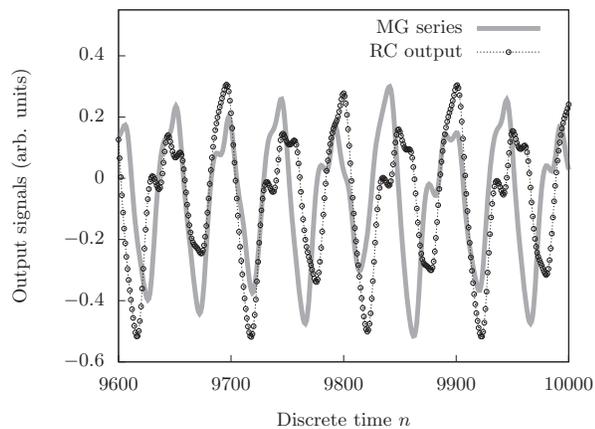}
  \caption{Output of the experimental reservoir computer (dotted black line) at the end of a long run of 10k timesteps. Although the system does not follow the starting trajectory (solid grey line), its output still resembles visually the target time series.}
  \label{fig:mglong}
\end{figure}

Fig. \ref{fig:reslarge} shows an example of the reservoir output $y(n)$ (dotted black line) during the autonomous run. The target Mackey-Glass series is shown in grey. The MSE threshold was set to $10^{-3}$ and the reservoir computer predicted 435 correct values in this example.
\textcolor{rev1}{Fig. \ref{fig:mgerrevo} displays the evolution of the MSE recorded during the same autonomous run. The plotted error curve was averaged over 200-timestep intervals. It exceeds the $10^{-3}$ threshold within $n\in[500,600]$ and reaches a constant value of approximately $1.1\e{-1}$ after 2500 timesteps. At this point, the generated time series is completely off the target (see Fig. \ref{fig:mglong} for illustration).}

The noise inside the opto-electronic reservoir, discussed in section \ref{subsec:resnoise}, makes the outcome of an experiment inconsistent. That is, several repetitions of the experiment with same parameters may result in significantly different prediction lengths. 
In fact, the impact of noise varies from one trial to another. In some cases it does not disturb the system much. But in most cases it induces a significant error on the output value $y(n)$, so that the neural network deviates very quickly from the target trajectory.
To estimate the variability of the results, we performed 50 consecutive autonomous runs with the same readout weights and the same optimal experimental parameters. While the system produced several very good predictions (of order of $400$), most of the outcomes were rather poor, with an average prediction length of $63.7$ and a standard deviation of $65.2$.
We obtained similar behaviour with the noisy experimental model, using the same level of noise as in the experiments.
Changing the ridge regression parameter in the training process (see section \ref{sec:rc}) did not improve the results.
This suggests that the reservoir computer emulates a ``noisy'' Mackey-Glass system, and therefore, using it to follow a given trajectory doesn't make much sense with such a high noise level.
Nevertheless, the noise does not prevent the system from emulating the Mackey-Glass system -- even if the output quickly deviates from the target, it still resembles the original time series.
Therefore, we tried a few distinct methods of comparing the output of the system with the target time series.

We performed a new set of experiments, where, after a training phase of $1500$ timesteps, the system was running autonomously for 10k timesteps \revs{in order} to collect enough points for data analysis.
We then proceeded with a simple visual inspection of the generated time series, to check whether it still looks similar to the Mackey-Glass time series, and does not settle down to simple periodic oscillations. Fig. \ref{fig:mglong} shows the output of the experimental reservoir computer at the end of the 10k-timestep autonomous run. It shows that the reservoir output is still similar to the target time series, that is, irregular and consisting of the same kind of uneven oscillations. 

\begin{figure}
  \includegraphics[width=0.45\textwidth]{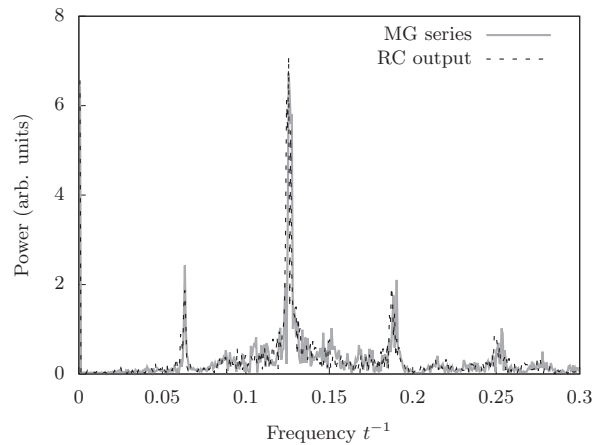}
  \caption{Comparison of Fast Fourier Transforms of the original Mackey-Glass series (solid grey line) and the time series generated by the photonic reservoir computer (dashed black line). The plot is limited to low frequencies as the power at higher frequencies is almost null. Dominant frequencies correspond to multiples of $1/\tau \approx 0.06$ (see section \ref{subsec:taskmg}). The experiment reproduces the target spectrum notably well.}
  \label{fig:fftmg}
\end{figure}

A more thorough way of comparing two time series that ``look similar'' is to compare their frequency spectra. Fig. \ref{fig:fftmg} shows the Fast Fourier Transforms of the original Mackey-Glass series (solid grey line) and the output of the experiment after a long run (dotted black line). Remarkably, the reservoir computer reproduces very accurately the spectrum of the chaotic time series, with its main frequency and several secondary frequencies. 

Finally, we estimated the Lyapunov exponent of the generated time series, using the method described in the Supplementary Material of \cite{jaeger2004harnessing}. We obtained $0.01$ for our experimental implementation, while the value commonly found in the literature for the Mackey-Glass series is $0.006$. The slightly higher value of the Lyapunov exponent may simply reflect the presence of noise in the emulator.

\subsection{Lorenz series prediction}

This task was investigated in a similar way to the previous one. The reservoir computer was trained over 3000 input samples and was run autonomously for 1000 timesteps. The 4000 samples were taken from an interval with even distribution of transitions between the two ``wings'' of the Lorenz attractor. In fact, we have noticed that the first 1000 samples of the sequence generated by the \verb+ode45+ solver (see section \ref{subsec:tasklz}) contained more oscillations above zero than below, that is, a transient from the starting point to the actual chaotic attractor. This uneven distribution forced the reservoir computer to generate a biased output. We thus discarded the first 1000 values and trained the system over the interval $\left[ 1000, 4000 \right]$  \textcolor{rev1}{(these initial transients were also removed in \cite{jaeger2004harnessing}).}
For optimal performance of the opto-electronic reservoir, we set the input gain to $\beta=0.5$ and the feedback attenuation to $\alpha = 6.1 \units{dB}$.

\begin{figure}
  \centering
  \includegraphics[width=0.45\textwidth]{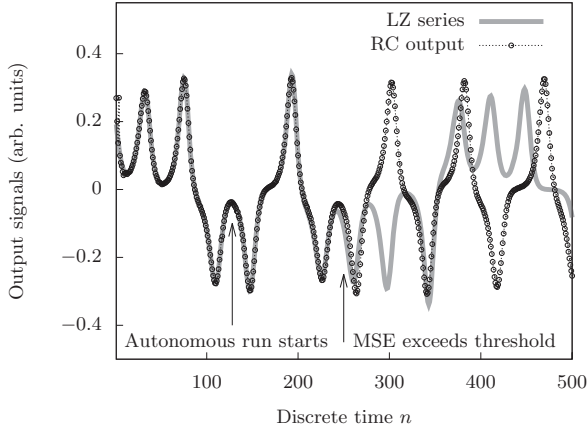}
  \caption{
    Example of reservoir computer output signal $y(n)$ (dotted black line) during autonomous runs on the Lorenz task. The system was driven by the target signal (solid grey line) for 128 timesteps before running autonomously (see section \ref{subsec:oeres}). The MSE threshold was set to $10^{-3}$. The photonic system with $N=600$ generated 122 correct values in this example, and predicted two switches of the trajectory from one lobe of the attractor to the other.
  }
  \label{fig:lzpred}
\end{figure}

\begin{figure}
  \centering
  \includegraphics[width=0.45\textwidth]{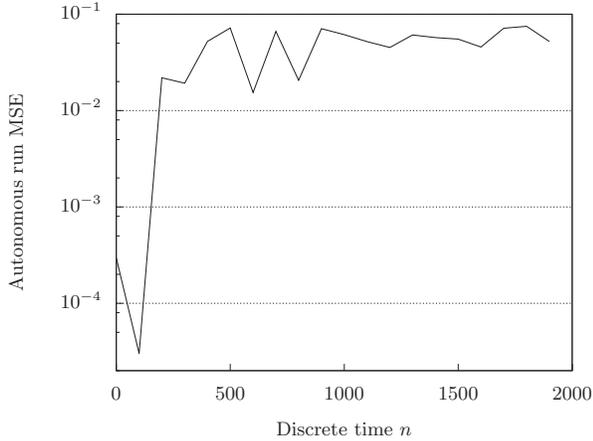}
  \caption{\textcolor{rev1}{
    Evolution of MSE during experimental autonomous generation of the Lorenz chaotic time series \revs{(same run as in Fig. (\ref{fig:lzpred})}. The error curve, averaged over 100 timesteps, crosses the $10^{-3}$ threshold near $n=250$. The initial dip corresponds to the warmup of the reservoir (see Sec. \ref{subsec:oeres}).
   }}
  \label{fig:lzerrevo}
\end{figure}

\begin{figure}
  \centering
  \includegraphics[width=0.45\textwidth]{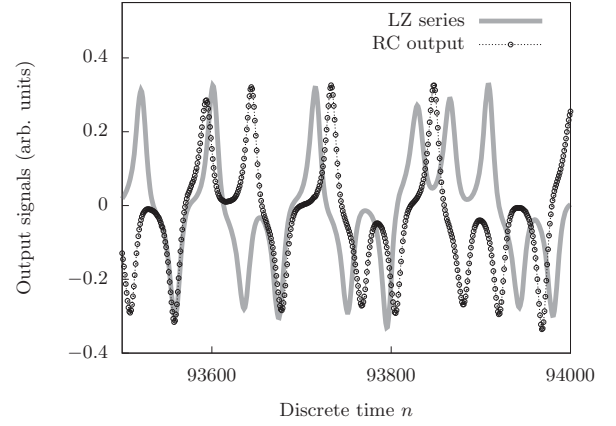}
  \caption{Output of the experiment (dotted black line) at the end of a long run of 95k timesteps on the Lorenz task. Although the system does not follow the starting trajectory (solid grey line), it does a fairly good job at emulating the dynamics of the Lorenz system.}
  \label{fig:lzend}
\end{figure}

Fig. \ref{fig:lzpred} shows an example of the reservoir output $y(n)$ (dotted black line) during the autonomous run. The target Lorenz series is shown in grey. With the MSE threshold set to $10^{-3}$, the system predicted 122 correct steps, including two transitions between the wings of the attractor. \revs{As} in the Mackey-Glass study, we performed 50 autonomous runs with identical readout weights and same optimal parameters, and obtained an average prediction horizon of $46.0$ timesteps with a standard deviation of $19.5$. Taking into account the higher degree of chaos of the Lorenz attractor, and given the same problems related to noise, it is hard to expect a better performance of the reservoir computer at following the target trajectory.
\textcolor{rev1}{Fig. \ref{fig:lzerrevo} depicts the evolution of the MSE during the autonomous run. The error curve was averaged over 100-timestep intervals. The initial dip corresonds to the teacher-forcing of the reservoir computer with the target signal for 128 timesteps, as discussed in Sec. \ref{subsec:oeres}. The error exceeds the $10^{-3}$ threshold around the $n=250$ mark and reaches a constant value of approximately $1.5\e{-2}$ after less than 1000 timesteps. At this point, the  reservoir computer has lost the target trajectory, but keeps on generating a time series with properties similar to the Lorenz series (see Fig. \ref{fig:lzend} for illustration).}

\begin{figure}
  \centering
  \includegraphics[width=0.45\textwidth]{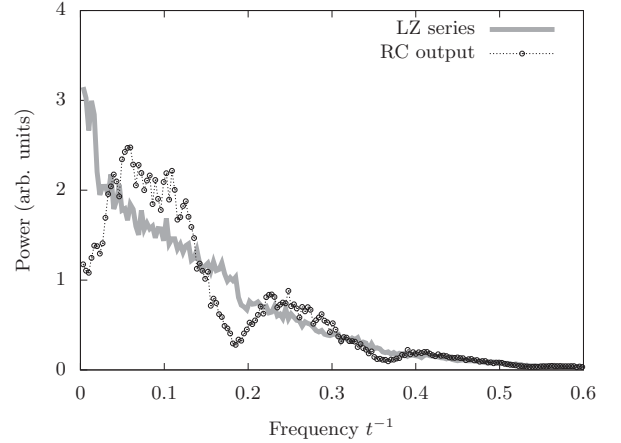}
  \caption{Comparison of Fast Fourier Transforms of the Lorenz series (solid grey line) and the time series generated by the photonic reservoir computer (dotted black line) during 95k timesteps. Both spectra are normalised so as to have equal total power. The curves are smoothened by averaging over 50 samples and the plot is limited to lower frequencies (the higher ones being close to zero). Despite some mismatch, the shape of the dotted curve is roughly similar to the grey line.}
  \label{fig:lzfft}
\end{figure}

Similar to the Mackey-Glass task, we performed a visual inspection of the generated Lorenz series after a long run, and compared the frequency spectra. Fig. \ref{fig:lzend} shows the output of the experiment near the end of a 95k autonomous run. Although the system is quite far from the target trajectory (plotted in grey) at this point, it is apparent that it has captured the dynamics of the Lorenz system very well. Fig. \ref{fig:lzfft} displays the Fast Fourier Transforms of the generated time series (dotted black line) and the computed Lorenz series (solid grey line). Unlike the Mackey-Glass system, these frequency spectra do not have any dominant frequencies. \textcolor{rev1}{That is, the power distribution does not contain any strong spikes, that could have been used as reference points for comparison. Therefore, comparing the two the spectra is much more subjective in this case. Although the curves do not match, one can still see a certain similitude between them.}

In addition to those visual comparisons, we performed a specific randomness test of the generated series. We exploited an interesting property of the Lorenz dynamics. Since it basically switches between two regions (the wings of the butterfly), with random transitions from one to the  other, one can assign binary ``0'' and ``1'' to these regions and thus transform the Lorenz series into a sequence of random bits. We used this trick to check the randomness of the generated series. We solved the Lorenz equation and ran the experiment for 95k timesteps and converted the resulting time series into two sequences of approximately 2400 bits. 
The two were then analysed with the ENT program \cite{ent} -- a well known software for testing random number sequences -- with the results shown in Tab. \ref{tab:resent}. Their interpretation requires a brief introduction of the tests performed by the software.

\begin{itemize}
  \item The first test computes the entropy per byte (8 bits). Since entropy is a measure of disorder, i.e. randomness, a truly random sequence should have 8 bits of entropy per byte. Both sequences are close to the maximum value, with the Lorenz series being slightly more random. 
  \item The compression is a commonly used indirect method of estimating the randomness of bytes in a file by compressing it with an efficient compression algorithm (such as e.g. Lempel-Ziv-Renau algorithm, used by the Zip program). These algorithms basically look for large repeating blocks, that should not appear in a truly random sequence. Again, both sequences could only be slightly compressed. 
  \item The mean value is the arithmetic mean of the data bytes. A random sequence should be evenly distributed, and thus have a random value of $127.5$. The Lorenz series is very close to this value, and the RC sequence is fairly close. 
  \item The Monte Carlo method of computing the value of $\pi$ randomly places points inside a square and computes the ratio of points located inside an inscribed circle, that is proportional to $\pi$. This is a more complex test that requires a large sequence to yield accurate results. We note that, nevertheless, both sequences produce a reasonable estimation of $\pi$.
  \item Finally, the serial correlation coefficient of a truly random sequence is zero. Both series present a very low correlation, yet again the Lorenz series demonstrating a better score.
\end{itemize}

These results do not prove that the generated sequence is random. One obviously has to use a much longer sequence of bits for this study, and should also consider more sophisticated and complete tests, such as Diehard \cite{diehard} or NIST Statistical Test Suite \cite{rukhin2001statistical}. The purpose of these tests was to show that the output of the RC generator does not consist of trivial oscillations, that only  remotly resemble the Lorenz system. The results show that the randomness of the RC output is similar to the Lorenz system, which gives reasons to believe in the similarity between the two time series. This, in turn, indicates, that our photonic reservoir computer was capable of learning to effectively emulate the dynamics of the Lorenz chaotic system.

\begin{table}
  \centering
  \begin{tabular}{|l|c|c|}
    \hline
    & RC & Lorenz \\
    \hline
    Entropy (/byte) & $6.6$ & $7.1$ \\
    Compression (\%) & 17 & 10 \\
    Mean (byte) & $134.3$ & $125.8$ \\
    $\pi$ & $2.88$ & $3.00$  \\
    Correlation & $-0.08$ & $-0.02$ \\
    \hline
  \end{tabular}
  \caption{Results returned by the ENT program for the bit sequences generated by the experiment (RC) and the integrated Lorenz system. The Lorenz sequence shows better figures, but the RC output is not far behind. All these figures are poor compared to common random series, but this is due to the very short sequences used here (roughly 300 bytes).}
  \label{tab:resent}
\end{table}

\section{Conclusion}
\label{sec:dis}

The present work demonstrates the potential of output feedback in hardware reservoir computing and constitutes an important step towards autonomous photonic implementations of recurrent neural networks. 
Specifically, we presented a photonic reservoir computer that is capable of generating both sine waves of different frequencies and short random patterns with substantial stability. Moreover, it could emulate the Mackey-Glass time series with a remarkably similar frequency spectrum, and a fairly close highest Lyapunov exponent. Finally, it could efficiently capture the dynamics of the Lorenz system and generate a time series with similar randomness properties. To the best of our knowledge, this is the first report of these task being implemented on an experimental reservoir computer.

The readout of the reservoir computer is carried out in real time on a fast FPGA chip. This results in a digital output layer in an analogue device, that nevertheless allows one to investigate many of the issues that will affect a system with purely analogue feedback. The latter is a much more complicated experiment. Indeed the only analogue output layers implemented so far on experimental reservoir computers were reported in \cite{smerieri2012analog,duport2016fully,vinckier2016autonomous}. Using them for output feedback would require adding an additional electronic circuit consisting of a sample and hold circuit, amplification, and multiplication  by the input mask. The present experiment allows one to investigate the benefits that output feedback has to offer to experimental reservoir computing, while anticipating the difficulties and limitations that will affect a fully analogue implementation. Such a two-step procedure, in which part of the experiment is analogue and part digital, is natural, and parallels the development of experimental reservoir computers in which some of the first experiments were only partially analogue, see e.g. \cite{appeltant2011information,martinenghi2012photonic}.

This work allowed to highlight a critical limitation of the present opto-electronic setup, \revs{namely} the relatively high level of noise generated by the components. While this was not a concern in previous experiments with offline readout \cite{paquot2012optoelectronic,duport2016fully}, it becomes critical in this study of the output feedback. Since this noise can not be averaged out, it propagates back into the system with output feedback and considerably deteriorates the reservoir states. 
This problem does not have a simple solution. One could rethink the entire experimental setup and rebuild it with new, less noisy components. One could also switch to a different experimental system, such as the low-noise passive cavity reported in \cite{vinckier2015high}. In any case, it will probably be difficult to increase the SNR above $60\units{dB}$. There may also be algorithmic solutions, such as using conceptors \cite{jaeger2014conceptors,jaeger2014controlling}. 

The high level of experimental noise quickly pushed the Reservoir Computer, emulating a chaotic system, away from a given trajectory, which initiated the search for quantitative methods for the evaluation of the experiment performance. In this work, we introduced a few simple techniques, \textcolor{rev1}{based on standard signal analysis methods,} such as statistics of the prediction length and visual comparison of the time series and their frequency spectra after a long autonomous run. We have also proposed case-specific methods, such as the randomness test, that could only be applied to the Lorenz time series. Overall, these are only the first steps towards the answer to a very general question: given a noisy emulator of a known chaotic system, how best to evaluate its performance? It will be interesting to understand the relationship between the performance obtained on the estimators above, and the properties of the chaotic system, such as the Lyapunov exponents or the dimension and geometry of the chaotic attractor. These questions should lead to a rich new direction of enquiry in the theory of nonlinear dynamics and complex systems.

\revs{Adding output feedback to experimental reservoir computers allows them to solve considerably more complex tasks than without output feedback. Future work could address nonlinear computations that depend on past information and that require persistent memory \cite{kovac2016persistent}, FORCE training \cite{sussillo2009generating} (which however requires that the learning time scale be short compared to the reservoir time scale) and applications such as frequency modulation  \cite{wyffels2014frequency}, or implementation of conceptors \cite{jaeger2014conceptors,jaeger2014controlling}. Ideally, an entirely analogue feedback should be implemented, like in e.g. \cite{antonik2017online}, rather than the digital feedback demonstrated here. The present work is therefore just a first step towards realising these additional applications.}

Finally, going back to the question of biological implementation, our work shows that the biologically plausible structure of reservoir computing \cite{maass2002real,yamazaki2007cerebellum,rossert2015edge} can be trained to generate highly complex temporal patterns, both periodic and chaotic, even in the presence of moderate levels of noise. Whether nature in fact implements this mechanism remains to be seen, and will depend amongst other aspects on the amount of noise present in biological implementations of reservoir computing, and whether there exist biologically plausible training mechanisms for this kind of signal generation.

We thank Michiel Hermans for insightful discussions. This work was supported by the Interuniversity Attraction Poles Program (Belgian Science Policy) Project Photonics@be IAP P7-35, by the Fonds de la Recherche Scientifique (FRS-FNRS) and by the Action de Recherche Concert\'ee of the F\'ed\'eration Universitaire Wallonie-Bruxelles through Grant No. AUWB-2012-12/17-ULB9.


%


\begin{thebibliography}{48}%
\makeatletter
\providecommand \@ifxundefined [1]{%
 \@ifx{#1\undefined}
}%
\providecommand \@ifnum [1]{%
 \ifnum #1\expandafter \@firstoftwo
 \else \expandafter \@secondoftwo
 \fi
}%
\providecommand \@ifx [1]{%
 \ifx #1\expandafter \@firstoftwo
 \else \expandafter \@secondoftwo
 \fi
}%
\providecommand \natexlab [1]{#1}%
\providecommand \enquote  [1]{``#1''}%
\providecommand \bibnamefont  [1]{#1}%
\providecommand \bibfnamefont [1]{#1}%
\providecommand \citenamefont [1]{#1}%
\providecommand \href@noop [0]{\@secondoftwo}%
\providecommand \href [0]{\begingroup \@sanitize@url \@href}%
\providecommand \@href[1]{\@@startlink{#1}\@@href}%
\providecommand \@@href[1]{\endgroup#1\@@endlink}%
\providecommand \@sanitize@url [0]{\catcode `\\12\catcode `\$12\catcode
  `\&12\catcode `\#12\catcode `\^12\catcode `\_12\catcode `\%12\relax}%
\providecommand \@@startlink[1]{}%
\providecommand \@@endlink[0]{}%
\providecommand \url  [0]{\begingroup\@sanitize@url \@url }%
\providecommand \@url [1]{\endgroup\@href {#1}{\urlprefix }}%
\providecommand \urlprefix  [0]{URL }%
\providecommand \Eprint [0]{\href }%
\providecommand \doibase [0]{http://dx.doi.org/}%
\providecommand \selectlanguage [0]{\@gobble}%
\providecommand \bibinfo  [0]{\@secondoftwo}%
\providecommand \bibfield  [0]{\@secondoftwo}%
\providecommand \translation [1]{[#1]}%
\providecommand \BibitemOpen [0]{}%
\providecommand \bibitemStop [0]{}%
\providecommand \bibitemNoStop [0]{.\EOS\space}%
\providecommand \EOS [0]{\spacefactor3000\relax}%
\providecommand \BibitemShut  [1]{\csname bibitem#1\endcsname}%
\let\auto@bib@innerbib\@empty
\bibitem [{\citenamefont {Jaeger}\ and\ \citenamefont
  {Haas}(2004)}]{jaeger2004harnessing}%
  \BibitemOpen
  \bibfield  {author} {\bibinfo {author} {\bibfnamefont {Herbert}\ \bibnamefont
  {Jaeger}}\ and\ \bibinfo {author} {\bibfnamefont {Harald}\ \bibnamefont
  {Haas}},\ }\bibfield  {title} {\enquote {\bibinfo {title} {Harnessing
  nonlinearity: Predicting chaotic systems and saving energy in wireless
  communication},}\ }\href@noop {} {\bibfield  {journal} {\bibinfo  {journal}
  {Science}\ }\textbf {\bibinfo {volume} {304}},\ \bibinfo {pages} {78--80}
  (\bibinfo {year} {2004})}\BibitemShut {NoStop}%
\bibitem [{\citenamefont {Maass}\ \emph {et~al.}(2002)\citenamefont {Maass},
  \citenamefont {Natschl{\"a}ger},\ and\ \citenamefont
  {Markram}}]{maass2002real}%
  \BibitemOpen
  \bibfield  {author} {\bibinfo {author} {\bibfnamefont {Wolfgang}\
  \bibnamefont {Maass}}, \bibinfo {author} {\bibfnamefont {Thomas}\
  \bibnamefont {Natschl{\"a}ger}}, \ and\ \bibinfo {author} {\bibfnamefont
  {Henry}\ \bibnamefont {Markram}},\ }\bibfield  {title} {\enquote {\bibinfo
  {title} {Real-time computing without stable states: {A} new framework for
  neural computation based on perturbations},}\ }\href@noop {} {\bibfield
  {journal} {\bibinfo  {journal} {Neural comput.}\ }\textbf {\bibinfo {volume}
  {14}},\ \bibinfo {pages} {2531--2560} (\bibinfo {year} {2002})}\BibitemShut
  {NoStop}%
\bibitem [{\citenamefont {Luko{\v{s}}evi{\v{c}}ius}\ and\ \citenamefont
  {Jaeger}(2009)}]{lukovsevivcius2009survey}%
  \BibitemOpen
  \bibfield  {author} {\bibinfo {author} {\bibfnamefont {Mantas}\ \bibnamefont
  {Luko{\v{s}}evi{\v{c}}ius}}\ and\ \bibinfo {author} {\bibfnamefont {Herbert}\
  \bibnamefont {Jaeger}},\ }\bibfield  {title} {\enquote {\bibinfo {title}
  {Reservoir computing approaches to recurrent neural network training},}\
  }\href@noop {} {\bibfield  {journal} {\bibinfo  {journal} {Comp. Sci. Rev.}\
  }\textbf {\bibinfo {volume} {3}},\ \bibinfo {pages} {127--149} (\bibinfo
  {year} {2009})}\BibitemShut {NoStop}%
\bibitem [{\citenamefont {Triefenbach}\ \emph {et~al.}(2010)\citenamefont
  {Triefenbach}, \citenamefont {Jalalvand}, \citenamefont {Schrauwen},\ and\
  \citenamefont {Martens}}]{triefenbach2010phoneme}%
  \BibitemOpen
  \bibfield  {author} {\bibinfo {author} {\bibfnamefont {Fabian}\ \bibnamefont
  {Triefenbach}}, \bibinfo {author} {\bibfnamefont {Azarakhsh}\ \bibnamefont
  {Jalalvand}}, \bibinfo {author} {\bibfnamefont {Benjamin}\ \bibnamefont
  {Schrauwen}}, \ and\ \bibinfo {author} {\bibfnamefont {Jean-Pierre}\
  \bibnamefont {Martens}},\ }\bibfield  {title} {\enquote {\bibinfo {title}
  {Phoneme recognition with large hierarchical reservoirs},}\ }\href@noop {}
  {\bibfield  {journal} {\bibinfo  {journal} {Adv. Neural Inf. Process. Syst.}\
  }\textbf {\bibinfo {volume} {23}},\ \bibinfo {pages} {2307--2315} (\bibinfo
  {year} {2010})}\BibitemShut {NoStop}%
\bibitem [{NFC(2006)}]{NFC}%
  \BibitemOpen
  \href@noop {} {\enquote {\bibinfo {title} {The 2006/07 forecasting
  competition for neural networks \& computational intelligence},}\ }\bibinfo
  {howpublished} {\url{http://www.neural-forecasting-competition.com/NN3/}}
  (\bibinfo {year} {2006})\BibitemShut {NoStop}%
\bibitem [{\citenamefont {Arsenault}(2012)}]{arsenault2012optical}%
  \BibitemOpen
  \bibfield  {author} {\bibinfo {author} {\bibfnamefont {Henri}\ \bibnamefont
  {Arsenault}},\ }\href@noop {} {\emph {\bibinfo {title} {Optical processing
  and computing}}}\ (\bibinfo  {publisher} {Elsevier},\ \bibinfo {year}
  {2012})\BibitemShut {NoStop}%
\bibitem [{\citenamefont {Appeltant}\ \emph {et~al.}(2011)\citenamefont
  {Appeltant}, \citenamefont {Soriano}, \citenamefont {Van~der Sande},
  \citenamefont {Danckaert}, \citenamefont {Massar}, \citenamefont {Dambre},
  \citenamefont {Schrauwen}, \citenamefont {Mirasso},\ and\ \citenamefont
  {Fischer}}]{appeltant2011information}%
  \BibitemOpen
  \bibfield  {author} {\bibinfo {author} {\bibfnamefont {Lennert}\ \bibnamefont
  {Appeltant}}, \bibinfo {author} {\bibfnamefont {Miguel~Cornelles}\
  \bibnamefont {Soriano}}, \bibinfo {author} {\bibfnamefont {Guy}\ \bibnamefont
  {Van~der Sande}}, \bibinfo {author} {\bibfnamefont {Jan}\ \bibnamefont
  {Danckaert}}, \bibinfo {author} {\bibfnamefont {Serge}\ \bibnamefont
  {Massar}}, \bibinfo {author} {\bibfnamefont {Joni}\ \bibnamefont {Dambre}},
  \bibinfo {author} {\bibfnamefont {Benjamin}\ \bibnamefont {Schrauwen}},
  \bibinfo {author} {\bibfnamefont {Claudio~R}\ \bibnamefont {Mirasso}}, \ and\
  \bibinfo {author} {\bibfnamefont {Ingo}\ \bibnamefont {Fischer}},\ }\bibfield
   {title} {\enquote {\bibinfo {title} {Information processing using a single
  dynamical node as complex system},}\ }\href@noop {} {\bibfield  {journal}
  {\bibinfo  {journal} {Nat. Commun.}\ }\textbf {\bibinfo {volume} {2}},\
  \bibinfo {pages} {468} (\bibinfo {year} {2011})}\BibitemShut {NoStop}%
\bibitem [{\citenamefont {Paquot}\ \emph {et~al.}(2012)\citenamefont {Paquot},
  \citenamefont {Duport}, \citenamefont {Smerieri}, \citenamefont {Dambre},
  \citenamefont {Schrauwen}, \citenamefont {Haelterman},\ and\ \citenamefont
  {Massar}}]{paquot2012optoelectronic}%
  \BibitemOpen
  \bibfield  {author} {\bibinfo {author} {\bibfnamefont {Yvan}\ \bibnamefont
  {Paquot}}, \bibinfo {author} {\bibfnamefont {Francois}\ \bibnamefont
  {Duport}}, \bibinfo {author} {\bibfnamefont {Anteo}\ \bibnamefont
  {Smerieri}}, \bibinfo {author} {\bibfnamefont {Joni}\ \bibnamefont {Dambre}},
  \bibinfo {author} {\bibfnamefont {Benjamin}\ \bibnamefont {Schrauwen}},
  \bibinfo {author} {\bibfnamefont {Marc}\ \bibnamefont {Haelterman}}, \ and\
  \bibinfo {author} {\bibfnamefont {Serge}\ \bibnamefont {Massar}},\ }\bibfield
   {title} {\enquote {\bibinfo {title} {Optoelectronic reservoir computing},}\
  }\href@noop {} {\bibfield  {journal} {\bibinfo  {journal} {Sci. Rep.}\
  }\textbf {\bibinfo {volume} {2}},\ \bibinfo {pages} {287} (\bibinfo {year}
  {2012})}\BibitemShut {NoStop}%
\bibitem [{\citenamefont {Larger}\ \emph {et~al.}(2012)\citenamefont {Larger},
  \citenamefont {Soriano}, \citenamefont {Brunner}, \citenamefont {Appeltant},
  \citenamefont {Guti{\'e}rrez}, \citenamefont {Pesquera}, \citenamefont
  {Mirasso},\ and\ \citenamefont {Fischer}}]{larger2012photonic}%
  \BibitemOpen
  \bibfield  {author} {\bibinfo {author} {\bibfnamefont {Laurent}\ \bibnamefont
  {Larger}}, \bibinfo {author} {\bibfnamefont {MC}~\bibnamefont {Soriano}},
  \bibinfo {author} {\bibfnamefont {Daniel}\ \bibnamefont {Brunner}}, \bibinfo
  {author} {\bibfnamefont {L}~\bibnamefont {Appeltant}}, \bibinfo {author}
  {\bibfnamefont {Jose~M}\ \bibnamefont {Guti{\'e}rrez}}, \bibinfo {author}
  {\bibfnamefont {Luis}\ \bibnamefont {Pesquera}}, \bibinfo {author}
  {\bibfnamefont {Claudio~R}\ \bibnamefont {Mirasso}}, \ and\ \bibinfo {author}
  {\bibfnamefont {Ingo}\ \bibnamefont {Fischer}},\ }\bibfield  {title}
  {\enquote {\bibinfo {title} {Photonic information processing beyond {T}uring:
  an optoelectronic implementation of reservoir computing},}\ }\href@noop {}
  {\bibfield  {journal} {\bibinfo  {journal} {Opt. Express}\ }\textbf {\bibinfo
  {volume} {20}},\ \bibinfo {pages} {3241--3249} (\bibinfo {year}
  {2012})}\BibitemShut {NoStop}%
\bibitem [{\citenamefont {Martinenghi}\ \emph {et~al.}(2012)\citenamefont
  {Martinenghi}, \citenamefont {Rybalko}, \citenamefont {Jacquot},
  \citenamefont {Chembo},\ and\ \citenamefont
  {Larger}}]{martinenghi2012photonic}%
  \BibitemOpen
  \bibfield  {author} {\bibinfo {author} {\bibfnamefont {Romain}\ \bibnamefont
  {Martinenghi}}, \bibinfo {author} {\bibfnamefont {Sergei}\ \bibnamefont
  {Rybalko}}, \bibinfo {author} {\bibfnamefont {Maxime}\ \bibnamefont
  {Jacquot}}, \bibinfo {author} {\bibfnamefont {Yanne~Kouomou}\ \bibnamefont
  {Chembo}}, \ and\ \bibinfo {author} {\bibfnamefont {Laurent}\ \bibnamefont
  {Larger}},\ }\bibfield  {title} {\enquote {\bibinfo {title} {Photonic
  nonlinear transient computing with multiple-delay wavelength dynamics},}\
  }\href@noop {} {\bibfield  {journal} {\bibinfo  {journal} {Phys. Rev. Let.}\
  }\textbf {\bibinfo {volume} {108}},\ \bibinfo {pages} {244101} (\bibinfo
  {year} {2012})}\BibitemShut {NoStop}%
\bibitem [{\citenamefont {Larger}\ \emph {et~al.}(2017)\citenamefont {Larger},
  \citenamefont {Bayl\'on-Fuentes}, \citenamefont {Martinenghi}, \citenamefont
  {Udaltsov}, \citenamefont {Chembo},\ and\ \citenamefont
  {Jacquot}}]{larger2017high}%
  \BibitemOpen
  \bibfield  {author} {\bibinfo {author} {\bibfnamefont {Laurent}\ \bibnamefont
  {Larger}}, \bibinfo {author} {\bibfnamefont {Antonio}\ \bibnamefont
  {Bayl\'on-Fuentes}}, \bibinfo {author} {\bibfnamefont {Romain}\ \bibnamefont
  {Martinenghi}}, \bibinfo {author} {\bibfnamefont {Vladimir~S.}\ \bibnamefont
  {Udaltsov}}, \bibinfo {author} {\bibfnamefont {Yanne~K.}\ \bibnamefont
  {Chembo}}, \ and\ \bibinfo {author} {\bibfnamefont {Maxime}\ \bibnamefont
  {Jacquot}},\ }\bibfield  {title} {\enquote {\bibinfo {title} {High-speed
  photonic reservoir computing using a time-delay-based architecture: Million
  words per second classification},}\ }\href@noop {} {\bibfield  {journal}
  {\bibinfo  {journal} {Phys. Rev. X}\ }\textbf {\bibinfo {volume} {7}},\
  \bibinfo {pages} {011015} (\bibinfo {year} {2017})}\BibitemShut {NoStop}%
\bibitem [{\citenamefont {Duport}\ \emph {et~al.}(2012)\citenamefont {Duport},
  \citenamefont {Schneider}, \citenamefont {Smerieri}, \citenamefont
  {Haelterman},\ and\ \citenamefont {Massar}}]{duport2012all}%
  \BibitemOpen
  \bibfield  {author} {\bibinfo {author} {\bibfnamefont {Fran{\c{c}}ois}\
  \bibnamefont {Duport}}, \bibinfo {author} {\bibfnamefont {Bendix}\
  \bibnamefont {Schneider}}, \bibinfo {author} {\bibfnamefont {Anteo}\
  \bibnamefont {Smerieri}}, \bibinfo {author} {\bibfnamefont {Marc}\
  \bibnamefont {Haelterman}}, \ and\ \bibinfo {author} {\bibfnamefont {Serge}\
  \bibnamefont {Massar}},\ }\bibfield  {title} {\enquote {\bibinfo {title}
  {All-optical reservoir computing},}\ }\href@noop {} {\bibfield  {journal}
  {\bibinfo  {journal} {Opt. Express}\ }\textbf {\bibinfo {volume} {20}},\
  \bibinfo {pages} {22783--22795} (\bibinfo {year} {2012})}\BibitemShut
  {NoStop}%
\bibitem [{\citenamefont {Brunner}\ \emph {et~al.}(2012)\citenamefont
  {Brunner}, \citenamefont {Soriano}, \citenamefont {Mirasso},\ and\
  \citenamefont {Fischer}}]{brunner2012parallel}%
  \BibitemOpen
  \bibfield  {author} {\bibinfo {author} {\bibfnamefont {D}~\bibnamefont
  {Brunner}}, \bibinfo {author} {\bibfnamefont {M.~C.}\ \bibnamefont
  {Soriano}}, \bibinfo {author} {\bibfnamefont {C.~R.}\ \bibnamefont
  {Mirasso}}, \ and\ \bibinfo {author} {\bibfnamefont {I.}~\bibnamefont
  {Fischer}},\ }\bibfield  {title} {\enquote {\bibinfo {title} {Parallel
  photonic information processing at gigabyte per second data rates using
  transient states},}\ }\href@noop {} {\bibfield  {journal} {\bibinfo
  {journal} {Nat. Commun.}\ }\textbf {\bibinfo {volume} {4}},\ \bibinfo {pages}
  {1364} (\bibinfo {year} {2012})}\BibitemShut {NoStop}%
\bibitem [{\citenamefont {Vinckier}\ \emph {et~al.}(2015)\citenamefont
  {Vinckier}, \citenamefont {Duport}, \citenamefont {Smerieri}, \citenamefont
  {Vandoorne}, \citenamefont {Bienstman}, \citenamefont {Haelterman},\ and\
  \citenamefont {Massar}}]{vinckier2015high}%
  \BibitemOpen
  \bibfield  {author} {\bibinfo {author} {\bibfnamefont {Quentin}\ \bibnamefont
  {Vinckier}}, \bibinfo {author} {\bibfnamefont {Fran\c{c}ois}\ \bibnamefont
  {Duport}}, \bibinfo {author} {\bibfnamefont {Anteo}\ \bibnamefont
  {Smerieri}}, \bibinfo {author} {\bibfnamefont {Kristof}\ \bibnamefont
  {Vandoorne}}, \bibinfo {author} {\bibfnamefont {Peter}\ \bibnamefont
  {Bienstman}}, \bibinfo {author} {\bibfnamefont {Marc}\ \bibnamefont
  {Haelterman}}, \ and\ \bibinfo {author} {\bibfnamefont {Serge}\ \bibnamefont
  {Massar}},\ }\bibfield  {title} {\enquote {\bibinfo {title} {High-performance
  photonic reservoir computer based on a coherently driven passive cavity},}\
  }\href@noop {} {\bibfield  {journal} {\bibinfo  {journal} {Optica}\ }\textbf
  {\bibinfo {volume} {2}},\ \bibinfo {pages} {438--446} (\bibinfo {year}
  {2015})}\BibitemShut {NoStop}%
\bibitem [{\citenamefont {Vandoorne}\ \emph {et~al.}(2014)\citenamefont
  {Vandoorne}, \citenamefont {Mechet}, \citenamefont {Van~Vaerenbergh},
  \citenamefont {Fiers}, \citenamefont {Morthier}, \citenamefont {Verstraeten},
  \citenamefont {Schrauwen}, \citenamefont {Dambre},\ and\ \citenamefont
  {Bienstman}}]{vandoorne2014experimental}%
  \BibitemOpen
  \bibfield  {author} {\bibinfo {author} {\bibfnamefont {Kristof}\ \bibnamefont
  {Vandoorne}}, \bibinfo {author} {\bibfnamefont {Pauline}\ \bibnamefont
  {Mechet}}, \bibinfo {author} {\bibfnamefont {Thomas}\ \bibnamefont
  {Van~Vaerenbergh}}, \bibinfo {author} {\bibfnamefont {Martin}\ \bibnamefont
  {Fiers}}, \bibinfo {author} {\bibfnamefont {Geert}\ \bibnamefont {Morthier}},
  \bibinfo {author} {\bibfnamefont {David}\ \bibnamefont {Verstraeten}},
  \bibinfo {author} {\bibfnamefont {Benjamin}\ \bibnamefont {Schrauwen}},
  \bibinfo {author} {\bibfnamefont {Joni}\ \bibnamefont {Dambre}}, \ and\
  \bibinfo {author} {\bibfnamefont {Peter}\ \bibnamefont {Bienstman}},\
  }\bibfield  {title} {\enquote {\bibinfo {title} {Experimental demonstration
  of reservoir computing on a silicon photonics chip},}\ }\href@noop {}
  {\bibfield  {journal} {\bibinfo  {journal} {Nat. Commun.}\ }\textbf {\bibinfo
  {volume} {5}},\ \bibinfo {pages} {3541} (\bibinfo {year} {2014})}\BibitemShut
  {NoStop}%
\bibitem [{\citenamefont {Zhang}(2012)}]{zhang2012neural}%
  \BibitemOpen
  \bibfield  {author} {\bibinfo {author} {\bibfnamefont {G.~Peter}\
  \bibnamefont {Zhang}},\ }\enquote {\bibinfo {title} {Neural networks for
  time-series forecasting},}\ in\ \href@noop {} {\emph {\bibinfo {booktitle}
  {Handbook of Natural Computing}}},\ \bibinfo {editor} {edited by\ \bibinfo
  {editor} {\bibfnamefont {Grzegorz}\ \bibnamefont {Rozenberg}}, \bibinfo
  {editor} {\bibfnamefont {Thomas}\ \bibnamefont {B{\"a}ck}}, \ and\ \bibinfo
  {editor} {\bibfnamefont {Joost~N.}\ \bibnamefont {Kok}}}\ (\bibinfo
  {publisher} {Springer Berlin Heidelberg},\ \bibinfo {address} {Berlin,
  Heidelberg},\ \bibinfo {year} {2012})\ pp.\ \bibinfo {pages}
  {461--477}\BibitemShut {NoStop}%
\bibitem [{\citenamefont {wyffels}\ and\ \citenamefont
  {Schrauwen}(2010)}]{wyffels2010comparative}%
  \BibitemOpen
  \bibfield  {author} {\bibinfo {author} {\bibfnamefont {F.}~\bibnamefont
  {wyffels}}\ and\ \bibinfo {author} {\bibfnamefont {B.}~\bibnamefont
  {Schrauwen}},\ }\bibfield  {title} {\enquote {\bibinfo {title} {A comparative
  study of reservoir computing strategies for monthly time series
  prediction},}\ }\href@noop {} {\bibfield  {journal} {\bibinfo  {journal}
  {Neurocomputing}\ }\textbf {\bibinfo {volume} {73}},\ \bibinfo {pages} {1958
  -- 1964} (\bibinfo {year} {2010})}\BibitemShut {NoStop}%
\bibitem [{\citenamefont {Antonik}\ \emph
  {et~al.}(2016{\natexlab{a}})\citenamefont {Antonik}, \citenamefont {Hermans},
  \citenamefont {Duport}, \citenamefont {Haelterman},\ and\ \citenamefont
  {Massar}}]{antonik2016towards}%
  \BibitemOpen
  \bibfield  {author} {\bibinfo {author} {\bibfnamefont {Piotr}\ \bibnamefont
  {Antonik}}, \bibinfo {author} {\bibfnamefont {Michiel}\ \bibnamefont
  {Hermans}}, \bibinfo {author} {\bibfnamefont {Fran{\c{c}}ois}\ \bibnamefont
  {Duport}}, \bibinfo {author} {\bibfnamefont {Marc}\ \bibnamefont
  {Haelterman}}, \ and\ \bibinfo {author} {\bibfnamefont {Serge}\ \bibnamefont
  {Massar}},\ }\bibfield  {title} {\enquote {\bibinfo {title} {Towards pattern
  generation and chaotic series prediction with photonic reservoir
  computers},}\ }in\ \href@noop {} {\emph {\bibinfo {booktitle} {SPIE's 2016
  Laser Technology and Industrial Laser Conference}}},\ Vol.\ \bibinfo {volume}
  {9732}\ (\bibinfo {year} {2016})\ p.\ \bibinfo {pages} {97320B}\BibitemShut
  {NoStop}%
\bibitem [{\citenamefont {Xu}\ \emph {et~al.}(2016)\citenamefont {Xu},
  \citenamefont {Han},\ and\ \citenamefont {Kanae}}]{xu2016lnorm}%
  \BibitemOpen
  \bibfield  {author} {\bibinfo {author} {\bibfnamefont {Meiling}\ \bibnamefont
  {Xu}}, \bibinfo {author} {\bibfnamefont {Min}\ \bibnamefont {Han}}, \ and\
  \bibinfo {author} {\bibfnamefont {Shunshoku}\ \bibnamefont {Kanae}},\
  }\bibfield  {title} {\enquote {\bibinfo {title} {L1/2 norm regularized echo
  state network for chaotic time series prediction},}\ }in\ \href@noop {}
  {\emph {\bibinfo {booktitle} {APNNS's 23th International Conference on Neural
  Information Processing}}},\ \bibinfo {series} {LNCS}, Vol.\ \bibinfo {volume}
  {9886}\ (\bibinfo {year} {2016})\ pp.\ \bibinfo {pages} {12--19}\BibitemShut
  {NoStop}%
\bibitem [{\citenamefont {wyffels}\ \emph {et~al.}(2008)\citenamefont
  {wyffels}, \citenamefont {Schrauwen},\ and\ \citenamefont
  {Stroobandt}}]{wyffels2008stable}%
  \BibitemOpen
  \bibfield  {author} {\bibinfo {author} {\bibfnamefont {Francis}\ \bibnamefont
  {wyffels}}, \bibinfo {author} {\bibfnamefont {Benjamin}\ \bibnamefont
  {Schrauwen}}, \ and\ \bibinfo {author} {\bibfnamefont {Dirk}\ \bibnamefont
  {Stroobandt}},\ }\bibfield  {title} {\enquote {\bibinfo {title} {Stable
  output feedback in reservoir computing using ridge regression},}\ }in\
  \href@noop {} {\emph {\bibinfo {booktitle} {International Conference on
  Artificial Neural Networks}}}\ (\bibinfo {organization} {Springer},\ \bibinfo
  {year} {2008})\ pp.\ \bibinfo {pages} {808--817}\BibitemShut {NoStop}%
\bibitem [{\citenamefont {Caluwaerts}\ \emph {et~al.}(2013)\citenamefont
  {Caluwaerts}, \citenamefont {D'Haene}, \citenamefont {Verstraeten},\ and\
  \citenamefont {Schrauwen}}]{caluwaerts2013locomotion}%
  \BibitemOpen
  \bibfield  {author} {\bibinfo {author} {\bibfnamefont {K.}~\bibnamefont
  {Caluwaerts}}, \bibinfo {author} {\bibfnamefont {M.}~\bibnamefont {D'Haene}},
  \bibinfo {author} {\bibfnamefont {D.}~\bibnamefont {Verstraeten}}, \ and\
  \bibinfo {author} {\bibfnamefont {B.}~\bibnamefont {Schrauwen}},\ }\bibfield
  {title} {\enquote {\bibinfo {title} {Locomotion without a brain: Physical
  reservoir computing in tensegrity structures},}\ }\href@noop {} {\bibfield
  {journal} {\bibinfo  {journal} {Artificial Life}\ }\textbf {\bibinfo {volume}
  {19}},\ \bibinfo {pages} {35 -- 66} (\bibinfo {year} {2013})}\BibitemShut
  {NoStop}%
\bibitem [{\citenamefont {Reinhart}\ and\ \citenamefont
  {Steil}(2012)}]{reinhart2012regularization}%
  \BibitemOpen
  \bibfield  {author} {\bibinfo {author} {\bibfnamefont {René~Felix}\
  \bibnamefont {Reinhart}}\ and\ \bibinfo {author} {\bibfnamefont
  {Jochen~Jakob}\ \bibnamefont {Steil}},\ }\bibfield  {title} {\enquote
  {\bibinfo {title} {Regularization and stability in reservoir networks with
  output feedback},}\ }\href@noop {} {\bibfield  {journal} {\bibinfo  {journal}
  {Neurocomputing}\ }\textbf {\bibinfo {volume} {90}},\ \bibinfo {pages} {96 --
  105} (\bibinfo {year} {2012})}\BibitemShut {NoStop}%
\bibitem [{\citenamefont {wyffels}\ \emph {et~al.}(2014)\citenamefont
  {wyffels}, \citenamefont {Li}, \citenamefont {Waegeman}, \citenamefont
  {Schrauwen},\ and\ \citenamefont {Jaeger}}]{wyffels2014frequency}%
  \BibitemOpen
  \bibfield  {author} {\bibinfo {author} {\bibfnamefont {Francis}\ \bibnamefont
  {wyffels}}, \bibinfo {author} {\bibfnamefont {Jiwen}\ \bibnamefont {Li}},
  \bibinfo {author} {\bibfnamefont {Tim}\ \bibnamefont {Waegeman}}, \bibinfo
  {author} {\bibfnamefont {Benjamin}\ \bibnamefont {Schrauwen}}, \ and\
  \bibinfo {author} {\bibfnamefont {Herbert}\ \bibnamefont {Jaeger}},\
  }\bibfield  {title} {\enquote {\bibinfo {title} {Frequency modulation of
  large oscillatory neural networks},}\ }\href@noop {} {\bibfield  {journal}
  {\bibinfo  {journal} {Biological Cybernetics}\ }\textbf {\bibinfo {volume}
  {108}},\ \bibinfo {pages} {145--157} (\bibinfo {year} {2014})}\BibitemShut
  {NoStop}%
\bibitem [{\citenamefont {Antonik}\ \emph
  {et~al.}(2016{\natexlab{b}})\citenamefont {Antonik}, \citenamefont {Hermans},
  \citenamefont {Haelterman},\ and\ \citenamefont
  {Massar}}]{antonik2016towards2}%
  \BibitemOpen
  \bibfield  {author} {\bibinfo {author} {\bibfnamefont {Piotr}\ \bibnamefont
  {Antonik}}, \bibinfo {author} {\bibfnamefont {Michiel}\ \bibnamefont
  {Hermans}}, \bibinfo {author} {\bibfnamefont {Marc}\ \bibnamefont
  {Haelterman}}, \ and\ \bibinfo {author} {\bibfnamefont {Serge}\ \bibnamefont
  {Massar}},\ }\bibfield  {title} {\enquote {\bibinfo {title} {Towards
  adjustable signal generation with photonic reservoir computers},}\ }in\
  \href@noop {} {\emph {\bibinfo {booktitle} {25th International Conference on
  Artificial Neural Networks}}},\ Vol.\ \bibinfo {volume} {9886}\ (\bibinfo
  {year} {2016})\BibitemShut {NoStop}%
\bibitem [{\citenamefont {Jaeger}(2007)}]{jaeger2007echo}%
  \BibitemOpen
  \bibfield  {author} {\bibinfo {author} {\bibfnamefont {H.}~\bibnamefont
  {Jaeger}},\ }\bibfield  {title} {\enquote {\bibinfo {title} {{E}cho state
  network},}\ }\href@noop {} {\bibfield  {journal} {\bibinfo  {journal}
  {Scholarpedia}\ }\textbf {\bibinfo {volume} {2}},\ \bibinfo {pages} {2330}
  (\bibinfo {year} {2007})}\BibitemShut {NoStop}%
\bibitem [{\citenamefont {Yamazaki}\ and\ \citenamefont
  {Tanaka}(2007)}]{yamazaki2007cerebellum}%
  \BibitemOpen
  \bibfield  {author} {\bibinfo {author} {\bibfnamefont {Tadashi}\ \bibnamefont
  {Yamazaki}}\ and\ \bibinfo {author} {\bibfnamefont {Shigeru}\ \bibnamefont
  {Tanaka}},\ }\bibfield  {title} {\enquote {\bibinfo {title} {The cerebellum
  as a liquid state machine},}\ }\href@noop {} {\bibfield  {journal} {\bibinfo
  {journal} {Neural Networks}\ }\textbf {\bibinfo {volume} {20}},\ \bibinfo
  {pages} {290 -- 297} (\bibinfo {year} {2007})}\BibitemShut {NoStop}%
\bibitem [{\citenamefont {R\"{o}ssert}\ \emph {et~al.}(2015)\citenamefont
  {R\"{o}ssert}, \citenamefont {Dean},\ and\ \citenamefont
  {Porrill}}]{rossert2015edge}%
  \BibitemOpen
  \bibfield  {author} {\bibinfo {author} {\bibfnamefont {Christian}\
  \bibnamefont {R\"{o}ssert}}, \bibinfo {author} {\bibfnamefont {Paul}\
  \bibnamefont {Dean}}, \ and\ \bibinfo {author} {\bibfnamefont {John}\
  \bibnamefont {Porrill}},\ }\bibfield  {title} {\enquote {\bibinfo {title} {At
  the edge of chaos: How cerebellar granular layer network dynamics can provide
  the basis for temporal filters},}\ }\href@noop {} {\bibfield  {journal}
  {\bibinfo  {journal} {PLOS Computational Biology}\ }\textbf {\bibinfo
  {volume} {11}},\ \bibinfo {pages} {1--28} (\bibinfo {year}
  {2015})}\BibitemShut {NoStop}%
\bibitem [{\citenamefont {Smerieri}\ \emph {et~al.}(2012)\citenamefont
  {Smerieri}, \citenamefont {Duport}, \citenamefont {Paquot}, \citenamefont
  {Schrauwen}, \citenamefont {Haelterman},\ and\ \citenamefont
  {Massar}}]{smerieri2012analog}%
  \BibitemOpen
  \bibfield  {author} {\bibinfo {author} {\bibfnamefont {Anteo}\ \bibnamefont
  {Smerieri}}, \bibinfo {author} {\bibfnamefont {Fran\c{c}ois}\ \bibnamefont
  {Duport}}, \bibinfo {author} {\bibfnamefont {Yvan}\ \bibnamefont {Paquot}},
  \bibinfo {author} {\bibfnamefont {Benjamin}\ \bibnamefont {Schrauwen}},
  \bibinfo {author} {\bibfnamefont {Marc}\ \bibnamefont {Haelterman}}, \ and\
  \bibinfo {author} {\bibfnamefont {Serge}\ \bibnamefont {Massar}},\ }\bibfield
   {title} {\enquote {\bibinfo {title} {Analog readout for optical reservoir
  computers},}\ \ }(\bibinfo {year} {2012})\ pp.\ \bibinfo {pages}
  {944--952}\BibitemShut {NoStop}%
\bibitem [{\citenamefont {Duport}\ \emph {et~al.}(2016)\citenamefont {Duport},
  \citenamefont {Smerieri}, \citenamefont {Akrout}, \citenamefont
  {Haelterman},\ and\ \citenamefont {Massar}}]{duport2016fully}%
  \BibitemOpen
  \bibfield  {author} {\bibinfo {author} {\bibfnamefont {Fran{\c{c}}ois}\
  \bibnamefont {Duport}}, \bibinfo {author} {\bibfnamefont {Anteo}\
  \bibnamefont {Smerieri}}, \bibinfo {author} {\bibfnamefont {Akram}\
  \bibnamefont {Akrout}}, \bibinfo {author} {\bibfnamefont {Marc}\ \bibnamefont
  {Haelterman}}, \ and\ \bibinfo {author} {\bibfnamefont {Serge}\ \bibnamefont
  {Massar}},\ }\bibfield  {title} {\enquote {\bibinfo {title} {Fully analogue
  photonic reservoir computer},}\ }\href@noop {} {\bibfield  {journal}
  {\bibinfo  {journal} {Sci. Rep.}\ }\textbf {\bibinfo {volume} {6}},\ \bibinfo
  {pages} {22381} (\bibinfo {year} {2016})}\BibitemShut {NoStop}%
\bibitem [{\citenamefont {Vinckier}\ \emph {et~al.}(2016)\citenamefont
  {Vinckier}, \citenamefont {Bouwens}, \citenamefont {Haelterman},\ and\
  \citenamefont {Massar}}]{vinckier2016autonomous}%
  \BibitemOpen
  \bibfield  {author} {\bibinfo {author} {\bibfnamefont {Quentin}\ \bibnamefont
  {Vinckier}}, \bibinfo {author} {\bibfnamefont {Arno}\ \bibnamefont
  {Bouwens}}, \bibinfo {author} {\bibfnamefont {Marc}\ \bibnamefont
  {Haelterman}}, \ and\ \bibinfo {author} {\bibfnamefont {Serge}\ \bibnamefont
  {Massar}},\ }\bibfield  {title} {\enquote {\bibinfo {title} {Autonomous
  all-photonic processor based on reservoir computing paradigm},}\ \ }(\bibinfo
   {publisher} {Optical Society of America},\ \bibinfo {year} {2016})\ p.\
  \bibinfo {pages} {SF1F.1}\BibitemShut {NoStop}%
\bibitem [{\citenamefont {Antonik}\ \emph
  {et~al.}(2016{\natexlab{c}})\citenamefont {Antonik}, \citenamefont {Duport},
  \citenamefont {Hermans}, \citenamefont {Smerieri}, \citenamefont
  {Haelterman},\ and\ \citenamefont {Massar}}]{antonik2016online}%
  \BibitemOpen
  \bibfield  {author} {\bibinfo {author} {\bibfnamefont {Piotr}\ \bibnamefont
  {Antonik}}, \bibinfo {author} {\bibfnamefont {François}\ \bibnamefont
  {Duport}}, \bibinfo {author} {\bibfnamefont {Michiel}\ \bibnamefont
  {Hermans}}, \bibinfo {author} {\bibfnamefont {Anteo}\ \bibnamefont
  {Smerieri}}, \bibinfo {author} {\bibfnamefont {Marc}\ \bibnamefont
  {Haelterman}}, \ and\ \bibinfo {author} {\bibfnamefont {Serge}\ \bibnamefont
  {Massar}},\ }\bibfield  {title} {\enquote {\bibinfo {title} {Online training
  of an opto-electronic reservoir computer applied to real-time channel
  equalization},}\ }\href@noop {} {\bibfield  {journal} {\bibinfo  {journal}
  {IEEE Transactions on Neural Networks and Learning Systems}\ }\textbf
  {\bibinfo {volume} {PP}},\ \bibinfo {pages} {1--13} (\bibinfo {year}
  {2016}{\natexlab{c}})}\BibitemShut {NoStop}%
\bibitem [{\citenamefont {Mackey}\ and\ \citenamefont
  {Glass}(1977)}]{mackey1977oscillation}%
  \BibitemOpen
  \bibfield  {author} {\bibinfo {author} {\bibfnamefont {Michael~C}\
  \bibnamefont {Mackey}}\ and\ \bibinfo {author} {\bibfnamefont {Leon}\
  \bibnamefont {Glass}},\ }\bibfield  {title} {\enquote {\bibinfo {title}
  {Oscillation and chaos in physiological control systems},}\ }\href@noop {}
  {\bibfield  {journal} {\bibinfo  {journal} {Science}\ }\textbf {\bibinfo
  {volume} {197}},\ \bibinfo {pages} {287--289} (\bibinfo {year}
  {1977})}\BibitemShut {NoStop}%
\bibitem [{\citenamefont {Lorenz}(1963)}]{lorenz1963deterministic}%
  \BibitemOpen
  \bibfield  {author} {\bibinfo {author} {\bibfnamefont {Edward~N}\
  \bibnamefont {Lorenz}},\ }\bibfield  {title} {\enquote {\bibinfo {title}
  {Deterministic nonperiodic flow},}\ }\href@noop {} {\bibfield  {journal}
  {\bibinfo  {journal} {Journal of the atmospheric sciences}\ }\textbf
  {\bibinfo {volume} {20}},\ \bibinfo {pages} {130--141} (\bibinfo {year}
  {1963})}\BibitemShut {NoStop}%
\bibitem [{\citenamefont {Rodan}\ and\ \citenamefont
  {Tino}(2011)}]{rodan2011minimum}%
  \BibitemOpen
  \bibfield  {author} {\bibinfo {author} {\bibfnamefont {Ali}\ \bibnamefont
  {Rodan}}\ and\ \bibinfo {author} {\bibfnamefont {Peter}\ \bibnamefont
  {Tino}},\ }\bibfield  {title} {\enquote {\bibinfo {title} {Minimum complexity
  echo state network},}\ }\href@noop {} {\bibfield  {journal} {\bibinfo
  {journal} {IEEE Trans. Neural Netw.}\ }\textbf {\bibinfo {volume} {22}},\
  \bibinfo {pages} {131--144} (\bibinfo {year} {2011})}\BibitemShut {NoStop}%
\bibitem [{\citenamefont {Tikhonov}\ \emph {et~al.}(1995)\citenamefont
  {Tikhonov}, \citenamefont {Goncharsky}, \citenamefont {Stepanov},\ and\
  \citenamefont {Yagola}}]{tikhonov1995numerical}%
  \BibitemOpen
  \bibfield  {author} {\bibinfo {author} {\bibfnamefont {Andrei~Nikolaevich}\
  \bibnamefont {Tikhonov}}, \bibinfo {author} {\bibfnamefont {AV}~\bibnamefont
  {Goncharsky}}, \bibinfo {author} {\bibfnamefont {VV}~\bibnamefont
  {Stepanov}}, \ and\ \bibinfo {author} {\bibfnamefont {Anatoly~G}\
  \bibnamefont {Yagola}},\ }\href@noop {} {\emph {\bibinfo {title} {Numerical
  methods for the solution of ill-posed problems}}},\ Vol.\ \bibinfo {volume}
  {328}\ (\bibinfo  {publisher} {Springer Netherlands},\ \bibinfo {year}
  {1995})\BibitemShut {NoStop}%
\bibitem [{\citenamefont {Farmer}(1982)}]{farmer1982chaotic}%
  \BibitemOpen
  \bibfield  {author} {\bibinfo {author} {\bibfnamefont {J~Doyne}\ \bibnamefont
  {Farmer}},\ }\bibfield  {title} {\enquote {\bibinfo {title} {Chaotic
  attractors of an infinite-dimensional dynamical system},}\ }\href@noop {}
  {\bibfield  {journal} {\bibinfo  {journal} {Physica D: Nonlinear Phenomena}\
  }\textbf {\bibinfo {volume} {4}},\ \bibinfo {pages} {366--393} (\bibinfo
  {year} {1982})}\BibitemShut {NoStop}%
\bibitem [{\citenamefont {Hirsch}\ \emph {et~al.}(2003)\citenamefont {Hirsch},
  \citenamefont {Smale},\ and\ \citenamefont
  {Devaney}}]{hirsch2003differential}%
  \BibitemOpen
  \bibfield  {author} {\bibinfo {author} {\bibfnamefont {Morris~W}\
  \bibnamefont {Hirsch}}, \bibinfo {author} {\bibfnamefont {Stephen}\
  \bibnamefont {Smale}}, \ and\ \bibinfo {author} {\bibfnamefont {Robert~L}\
  \bibnamefont {Devaney}},\ }\href@noop {} {\emph {\bibinfo {title}
  {Differential equations, dynamical systems, and an introduction to chaos}}}\
  (\bibinfo  {publisher} {Academic press, Boston, MA},\ \bibinfo {year}
  {2003})\BibitemShut {NoStop}%
\bibitem [{iee(1994)}]{ieeevhdl}%
  \BibitemOpen
  \bibfield  {title} {\enquote {\bibinfo {title} {{IEEE} {S}tandard {VHDL}
  {L}anguage {R}eference {M}anual.}}\ }\href {\doibase
  10.1109/IEEESTD.1994.121433} {\bibfield  {journal} {\bibinfo  {journal}
  {ANSI/IEEE Std 1076-1993}\ } (\bibinfo {year} {1994}),\
  10.1109/IEEESTD.1994.121433}\BibitemShut {NoStop}%
\bibitem [{\citenamefont {Pedroni}(2004)}]{pedroni2004circuit}%
  \BibitemOpen
  \bibfield  {author} {\bibinfo {author} {\bibfnamefont {V.A.}\ \bibnamefont
  {Pedroni}},\ }\href@noop {} {\emph {\bibinfo {title} {Circuit Design with
  VHDL}}}\ (\bibinfo  {publisher} {MIT Press},\ \bibinfo {year}
  {2004})\BibitemShut {NoStop}%
\bibitem [{\citenamefont {Horowitz}\ and\ \citenamefont
  {Hill}(1980)}]{horowitz1980art}%
  \BibitemOpen
  \bibfield  {author} {\bibinfo {author} {\bibfnamefont {Paul}\ \bibnamefont
  {Horowitz}}\ and\ \bibinfo {author} {\bibfnamefont {Winfield}\ \bibnamefont
  {Hill}},\ }\href@noop {} {\emph {\bibinfo {title} {The art of electronics}}}\
  (\bibinfo  {publisher} {Cambridge University Press},\ \bibinfo {year}
  {1980})\BibitemShut {NoStop}%
\bibitem [{\citenamefont {Walker}()}]{ent}%
  \BibitemOpen
  \bibfield  {author} {\bibinfo {author} {\bibfnamefont {John}\ \bibnamefont
  {Walker}},\ }\href@noop {} {\enquote {\bibinfo {title} {{ENT} program},}\
  }\bibinfo {howpublished} {\url{http://www.fourmilab.ch/random/}}\BibitemShut
  {NoStop}%
\bibitem [{\citenamefont {Marsaglia}()}]{diehard}%
  \BibitemOpen
  \bibfield  {author} {\bibinfo {author} {\bibfnamefont {George}\ \bibnamefont
  {Marsaglia}},\ }\href@noop {} {\enquote {\bibinfo {title} {The {M}arsaglia
  {R}andom {N}umber {CDROM} including the {D}iehard {B}attery of {T}ests of
  {R}andomness},}\ }\bibinfo {howpublished}
  {\url{http://stat.fsu.edu/pub/diehard/}}\BibitemShut {NoStop}%
\bibitem [{\citenamefont {Rukhin}\ \emph {et~al.}(2001)\citenamefont {Rukhin},
  \citenamefont {Soto}, \citenamefont {Nechvatal}, \citenamefont {Smid},\ and\
  \citenamefont {Barker}}]{rukhin2001statistical}%
  \BibitemOpen
  \bibfield  {author} {\bibinfo {author} {\bibfnamefont {Andrew}\ \bibnamefont
  {Rukhin}}, \bibinfo {author} {\bibfnamefont {Juan}\ \bibnamefont {Soto}},
  \bibinfo {author} {\bibfnamefont {James}\ \bibnamefont {Nechvatal}}, \bibinfo
  {author} {\bibfnamefont {Miles}\ \bibnamefont {Smid}}, \ and\ \bibinfo
  {author} {\bibfnamefont {Elaine}\ \bibnamefont {Barker}},\ }\href@noop {}
  {\emph {\bibinfo {title} {A statistical test suite for random and
  pseudorandom number generators for cryptographic applications}}},\ \bibinfo
  {type} {Tech. Rep.}\ (\bibinfo  {institution} {National Institute of
  Standards and Technology},\ \bibinfo {year} {2001})\BibitemShut {NoStop}%
\bibitem [{\citenamefont {Jaeger}(2014{\natexlab{a}})}]{jaeger2014conceptors}%
  \BibitemOpen
  \bibfield  {author} {\bibinfo {author} {\bibfnamefont {Herbert}\ \bibnamefont
  {Jaeger}},\ }\bibfield  {title} {\enquote {\bibinfo {title} {Conceptors: an
  easy introduction},}\ }\href@noop {} {\bibfield  {journal} {\bibinfo
  {journal} {CoRR}\ }\textbf {\bibinfo {volume} {abs/1406.2671}} (\bibinfo
  {year} {2014}{\natexlab{a}})}\BibitemShut {NoStop}%
\bibitem [{\citenamefont {Jaeger}(2014{\natexlab{b}})}]{jaeger2014controlling}%
  \BibitemOpen
  \bibfield  {author} {\bibinfo {author} {\bibfnamefont {Herbert}\ \bibnamefont
  {Jaeger}},\ }\bibfield  {title} {\enquote {\bibinfo {title} {Controlling
  recurrent neural networks by conceptors},}\ }\href@noop {} {\bibfield
  {journal} {\bibinfo  {journal} {CoRR}\ }\textbf {\bibinfo {volume}
  {abs/1403.3369}} (\bibinfo {year} {2014}{\natexlab{b}})}\BibitemShut
  {NoStop}%
\bibitem [{\citenamefont {Kovac}\ \emph {et~al.}(2016)\citenamefont {Kovac},
  \citenamefont {Koall}, \citenamefont {Pipa},\ and\ \citenamefont
  {Toutounji}}]{kovac2016persistent}%
  \BibitemOpen
  \bibfield  {author} {\bibinfo {author} {\bibfnamefont {André~David}\
  \bibnamefont {Kovac}}, \bibinfo {author} {\bibfnamefont {Maximilian}\
  \bibnamefont {Koall}}, \bibinfo {author} {\bibfnamefont {Gordon}\
  \bibnamefont {Pipa}}, \ and\ \bibinfo {author} {\bibfnamefont {Hazem}\
  \bibnamefont {Toutounji}},\ }\bibfield  {title} {\enquote {\bibinfo {title}
  {Persistent memory in single node delay-coupled reservoir computing},}\
  }\href@noop {} {\bibfield  {journal} {\bibinfo  {journal} {PLOS ONE}\
  }\textbf {\bibinfo {volume} {11}},\ \bibinfo {pages} {1--15} (\bibinfo {year}
  {2016})}\BibitemShut {NoStop}%
\bibitem [{\citenamefont {Sussillo}\ and\ \citenamefont
  {Abbott}(2009)}]{sussillo2009generating}%
  \BibitemOpen
  \bibfield  {author} {\bibinfo {author} {\bibfnamefont {David}\ \bibnamefont
  {Sussillo}}\ and\ \bibinfo {author} {\bibfnamefont {L.F.}\ \bibnamefont
  {Abbott}},\ }\bibfield  {title} {\enquote {\bibinfo {title} {Generating
  coherent patterns of activity from chaotic neural networks},}\ }\href@noop {}
  {\bibfield  {journal} {\bibinfo  {journal} {Neuron}\ }\textbf {\bibinfo
  {volume} {63}},\ \bibinfo {pages} {544 -- 557} (\bibinfo {year}
  {2009})}\BibitemShut {NoStop}%
\bibitem [{\citenamefont {Antonik}\ \emph {et~al.}(2017)\citenamefont
  {Antonik}, \citenamefont {Haelterman},\ and\ \citenamefont
  {Massar}}]{antonik2017online}%
  \BibitemOpen
  \bibfield  {author} {\bibinfo {author} {\bibfnamefont {Piotr}\ \bibnamefont
  {Antonik}}, \bibinfo {author} {\bibfnamefont {Marc}\ \bibnamefont
  {Haelterman}}, \ and\ \bibinfo {author} {\bibfnamefont {Serge}\ \bibnamefont
  {Massar}},\ }\bibfield  {title} {\enquote {\bibinfo {title} {Online training
  for high-performance analogue readout layers in photonic reservoir
  computers},}\ }\href@noop {} {\bibfield  {journal} {\bibinfo  {journal}
  {Cognitive Computation}\ ,\ \bibinfo {pages} {1--10}} (\bibinfo {year}
  {2017})}\BibitemShut {NoStop}%
\end{thebibliography}
%
\end{document}